%% file: main.tex
\documentclass[twoside]{article}

\usepackage[accepted]{aistats/aistats2026}
\usepackage{graphicx}
\usepackage{hyperref}
\usepackage{subcaption}
\usepackage{amsfonts}
\usepackage{nicefrac}
\usepackage{xcolor}
\usepackage{wrapfig}
\usepackage[symbol]{footmisc}

\newcommand{\vx}{\ensuremath{\mathbf{x}}}
\newcommand{\norm}[1]{\left\lVert#1\right\rVert}

\newcommand{\method}{BOAT}

\usepackage[round]{natbib}

\begin{document}

%
\runningtitle{BOAT: Antibody Design via Multi-Objective Bayesian Optimization}

%
\runningauthor{Rao, Gonzalez, Gerard, Gessner}

\twocolumn[

\aistatstitle{BOAT: Navigating the Sea of In Silico Predictors for Antibody Design via Multi-Objective Bayesian Optimization}

\aistatsauthor{Jackie Rao$^1$ \And Ferran Gonzalez Hernandez$^2$ \And Leon Gerard$^2$ \And Alexandra Gessner$^2$}

\aistatsaddress{$^1$MRC Biostatistics Unit, University of Cambridge, UK \\  $^2$Centre for AI, Data Science and Artificial Intelligence, R\&D, AstraZeneca, Barcelona, Spain}
]
\begin{abstract}
    Antibody lead optimization is inherently a multi-objective challenge in drug discovery. Achieving a balance between different drug-like properties is crucial for the development of viable candidates, and this search becomes exponentially challenging as desired properties grow. The ever-growing zoo of sophisticated \textit{in silico} tools for predicting antibody properties calls for an efficient joint optimization procedure to overcome resource-intensive sequential filtering pipelines. We present \method, a versatile Bayesian optimization framework for multi-property antibody engineering. Our `plug-and-play' framework couples uncertainty-aware surrogate modeling with a genetic algorithm to jointly optimize various predicted antibody traits while enabling efficient exploration of sequence space. Through systematic benchmarking against genetic algorithms and newer generative learning approaches, we demonstrate competitive performance with state-of-the-art methods for multi-objective protein optimization. We identify clear regimes where surrogate-driven optimization outperforms expensive generative approaches and establish practical limits imposed by sequence dimensionality and oracle costs. 
\end{abstract}

\section{INTRODUCTION}

Lead optimization lies at the heart of therapeutic antibody development, where the goal is to advance promising candidates into clinically viable drugs. In this process, candidates are systematically improved to meet multiple, often competing, criteria such as binding affinity, manufacturability, biophysical stability, and immunogenicity. The specific combination of properties targeted can vary significantly between campaigns, with some requiring cross-species reactivity to enable testing human therapeutics in animal models, while others prioritize developability or other traits. Optimization efforts typically focus on the complementarity-determining regions (CDRs) of the antibody, the variable loops responsible for antigen binding \citep{sela2013structural}. Heavy chain CDRs are often prioritized, particularly CDR-H3, which exhibits the greatest diversity and contributes most significantly to binding \citep{xu2000diversity}. As the number of required properties grows, the complexity of searching for optimal antibody sequences quickly outpaces what can be achieved through traditional trial-and-error or single-target screening methods. The exponential growth in sequence and property space creates a pressing need for systematic strategies that can efficiently navigate these multidimensional landscapes.

Modern \textit{in silico} approaches have emerged as indispensable tools in addressing these challenges, allowing scientists to rapidly predict and evaluate protein features before experimental validation. Machine learning-based property predictors and physics-based simulations offer the potential to assess vast libraries of antibody variants at a much lower cost than experimental validation. Still, the computational resources required for tasks such as structure prediction or simulating relative binding free energies are substantial. Hence, a systematic approach is needed to decide which candidates to score. Furthermore, integrating these predictive models for multiple objectives presents methodological hurdles: conflicting property requirements may restrict sequence innovation, and poor predictive power complicates decision-making. Therefore, methodologies capable of jointly optimizing multiple objectives while quantifying uncertainty are vital for designing new experiments and steering antibody engineering towards the most promising leads. 

\begin{figure}[t]
\includegraphics[width=\columnwidth]{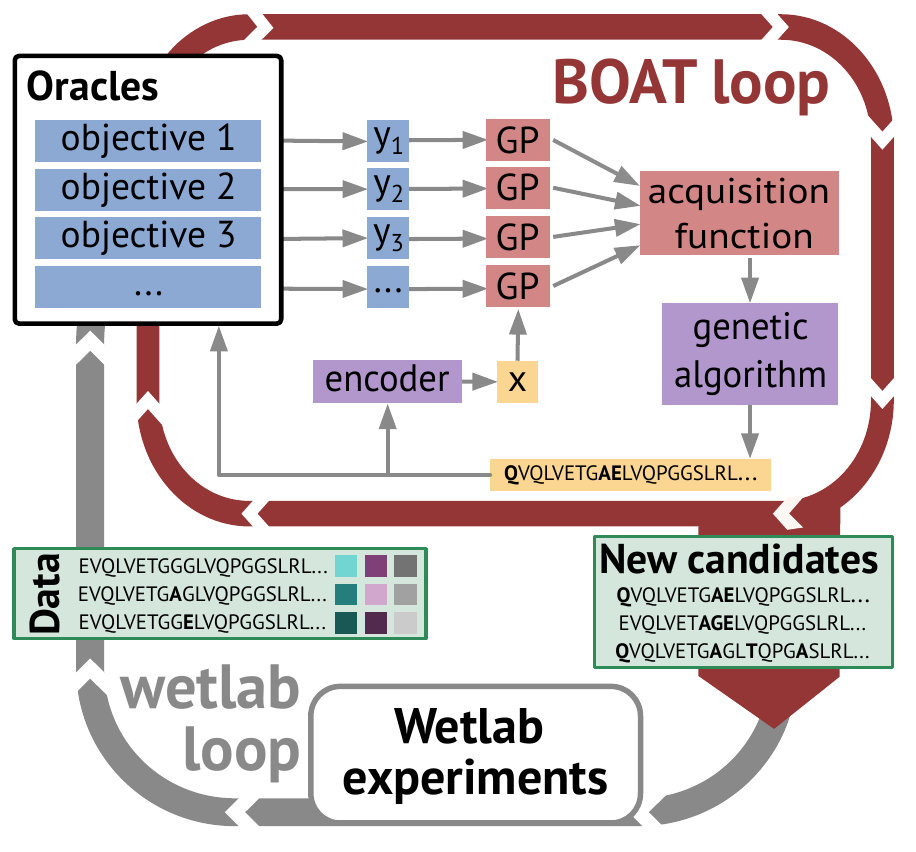}
\caption{Illustration of the \method~loop for lead optimization (LO). LO campaigns typically consist of a few rounds of experimental testing of the order of a hundred sequences in the wetlab, where previous experiments inform the sequence design for the next iteration of the wetlab loop. The \method~loop can be thought of as an inner loop in a single sequence design step of the wetlab loop, leveraging sophisticated tools used for computationally assessing sequences. The multi-objective optimization loop eliminates the common practice of generating a large pool of sequences and filtering them down by sequentially passing them through the different oracles, which is computationally inefficient and unaware of Pareto trade-offs. The oracles used depend on the requirements of the LO campaign, and can be exchanged flexibly. See Section~\ref{sec:methods} for details on the internal building blocks of \method.}
\label{fig:boat}
\end{figure}

We address these challenges with \method~(\textbf{B}ayesian \textbf{O}ptimization for \textbf{A}ntibody \textbf{T}raits), a versatile multi-objective Bayesian optimization framework for antibody sequences, illustrated in Figure~\ref{fig:boat}. \method~supports easy interfacing of arbitrary \textit{in silico} predictors, and allows for either full sequence optimization or region-specific optimization. This allows users leverage appropriate scoring functions depending on particular requirements posed in lead optimization campaigns.
\method~has been constructed as the inner loop in a single sequence design step of an outer "wetlab loop". While data from previous wetlab experiments may be used to inform oracles, \method~remains agnostic of experimental data. Thus, it remains up to the user to validate their selected oracles prior to optimizing them with \method.
While we focus here on antibodies, the approach extends naturally to other therapeutic proteins and small molecules. In this work, we consider models for binding affinity, humanness evaluation, as well as structure prediction for joint optimization. Humanness prediction assesses how closely an engineered antibody sequence resembles naturally occurring human antibodies—a critical property for therapeutic development, as higher humanness typically reduces immunogenicity and improves safety profiles.

\subsection*{Key Contributions}

\begin{itemize}
    \item We construct a light-weight `plug-and-play' Bayesian multi-objective optimization framework to optimize antibody lead candidates against computationally predicted properties of interest. Code can be found at \url{https://github.com/AstraZeneca/boat}.
    \item We perform rigorous benchmarking of Bayesian-based optimization with surrogate models versus genetic and generative baselines, quantifying oracle efficiency, diversity and Pareto front quality.
    \item We demonstrate that our method efficiently explores the Pareto front where the combinatorial ground truth is available. This approach enables the systematic identification of Pareto optimal candidates, allowing for the selection of antibodies that represent balanced trade-offs between multiple objectives according to specific priorities.
    \item We deliver guidelines for the selection of approaches for experimental design in antibody engineering, grounded in systematic scaling and integration of both inexpensive and computationally demanding oracles.
\end{itemize}

\section{MATERIALS AND METHODS}
\label{sec:methods}

\subsection{Multi-Objective Bayesian Optimization}

Bayesian Optimization (BO) provides a sample-efficient framework for global optimization by maintaining a probabilistic surrogate model of the objective function and using acquisition functions to guide the search. See \citet{frazier2018tutorial, garnett2023bayesian} for a detailed introduction to BO. Generally, BO seeks to find the maximum of a function $f: \mathcal{X} \rightarrow \mathbb{R}$ in as few evaluations as possible, where $f$ is often expensive to evaluate and lacks structure (e.g., closed-form gradients) that would make it amenable to direct optimization methods. 

Given a dataset of previous (potentially noisy) evaluations $\mathcal{D}_t = {(\mathbf{x}_i, y_i)}_{i = 1, ..., t}$, a probabilistic surrogate model $p(f|\mathcal{D}_t)$ is fit to this dataset which captures the current belief about the unknown objective function $f$, where $y_i = f(\mathbf{x}_i) + \epsilon_i$ and $\epsilon_i \sim \mathcal{N}(0, \sigma^2)$. A Gaussian process (GP) \citep{williams2006gaussian} is the most commonly used surrogate model, enabling non-parametric regression and providing both mean and uncertainty estimates at each input point $\mathbf{x}$. GPs are popular due to their closed-form posteriors, flexibility in kernel choice for encoding different data structures and inductive biases, and strong performance in data-scarce regimes. 

An acquisition function $\alpha(\mathbf{x}|\mathcal{D}_t)$ quantifies the utility of evaluating $f$ at each candidate input point, given predictions from the surrogate model, balancing exploitation of regions with high predictive performance against exploration of uncertain and under-explored areas. The next evaluation point is selected by optimizing this acquisition function: $\mathbf{x}_{t+1} = \arg\max_{\mathbf{x}} \alpha(\mathbf{x}|\mathcal{D}_t)$. After evaluating $y_{t+1} = f(\mathbf{x}_{t+1})$, the dataset is augmented and the process repeats until a stopping criterion is met or the evaluation budget is exhausted. For single-objective optimization, we use Log Expected Improvement (LogEI) \citep{ament2023unexpected}, and for multi-objective optimization, we use Expected Hypervolume Improvement (EHVI) \citep{emmerich2011hypervolume} and its noisy extension, Noisy Expected Hypervolume Improvement (NEHVI) \citep{daulton2021parallel}. These objective functions promote expansion of the Pareto front and maximization of the associated hypervolume. We note that other multi-objective alternatives exist, such as MORBO \citep{daulton2022multi} and ParEGO \citep{knowles2006parego}. Our implementation leverages the BOTorch framework \citep{balandat2020botorch}, which also allows batch extensions of the acquisition functions above \citep{daulton2020differentiable} and whose modular design enables straightforward extension to additional acquisition functions.

\subsection{BO in Sequence Space}\label{ref:boseq}

Common kernels for Gaussian processes map from $\mathbb{R}^d \times \mathbb{R}^d$ or a subset thereof to the real line. To apply Bayesian optimization to sequences of amino acids defined by strings $s\in \mathcal{S}$, there are two options, 1) to define a string kernel that operates on string space directly, or 2) to embed the sequences to represent them in a numerical space. We choose the latter approach and consider the following sequence encodings,
\begin{description}
    \item[One-hot] Each amino acid gets encoded as a one-hot vector; sequences are encoded as a concatenation of one-hot encoded amino acids.
    \item[Bag of amino acids] To include sequence motifs beyond individual amino acids, we encode matching $n$-grams (with $n$=5), similar to the bag of words embedding.
    \item[BLOSUM] We follow \citep{oglic2018learning, oglic2019scalable, gessner2024active} and use the eigendecomposition $UDU^T$ of the block-substitution matrix (BLOSUM) with similarity 45 to construct embedding vectors $U|D|^{\nicefrac{1}{2}}$. BLOSUM \citep{henikoff1992amino} is an indefinite matrix that quantifies similarities between amino acids by recording the effect of their substitution in proteins.  
    \item[AbLang-2] AbLang-2 is an antibody-specific protein language model providing embeddings of antibody sequences taking into account learned context across the sequence \citep{olsen2024addressing}.
\end{description}

The embedding space is typically quite large for all considered embeddings. For example, both one-hot and BLOSUM give rise to sequence embeddings of size sequence length $\times$ number of amino acids (i.e., 20). We employ a Gaussian process model that has been designed for this kind of high-dimensional problem, using the Tanimoto kernel \citep{ralaivola2005graph}
\begin{equation}
    k_\mathrm{Tanimoto} (\vx, \vx') = \frac{\langle\vx,\vx'\rangle}{\norm{\vx}^2 + \norm{\vx'}^2 - \langle\vx,\vx'\rangle}.
\end{equation}
The Tanimoto similarity was initially used to compare binary molecular fingerprints of small molecules, but has been extended for more general molecular embeddings that lie in $\mathbb{R}^d$.

\subsection{Genetic Optimizer} \label{sec:GA}

While some embeddings, such as one-hot and BLOSUM, admit an explicit reconstruction of amino acid sequences, they still represent fundamentally discrete objects. Although the embeddings are vector-valued, each dimension corresponds to categorical amino-acid choice, and most points in this vector space do not correspond to valid sequences. While the acquisition function is differentiable in principle, a gradient-based optimizer would move through arbitrary real-valued vectors; projecting these vectors back to the nearest valid discrete sequence would result in large, semantically meaningless jumps. We instead use a genetic algorithm (GA), a discrete optimization method, to generate sequences guided by the acquisition score.
In each iteration of the BO loop, we generate an initial population by slightly mutating the previously evaluated sequences -- this way, we ensure not to start in a local minimum of the acquisition function. To generate a new generation, we repeatedly apply:
\begin{description}
    \item[Tournament selection] Sample a subset of the previous generation and retain the best-scoring sequence.
    \item[Single-point crossover] Having sampled two parents via tournament selection, we create two offsprings by randomly cutting both parental sequences at a sampled position and swapping the remaining sequence after this position. We apply crossover with a rate of $0.7$, otherwise the parents make it to the next step.
    \item[Mutation] We then apply random mutations to amino acids in the sequence. We use a per-position mutation probability of either $0.1$ or $0.15$.
\end{description}
This procedure is repeated until the new generation has the desired size. The parameters have been chosen from an initial tuning phase. If not stated otherwise, we use an initial population of size $50$, and $50$ sequences per generation over $20$ generations. The score used is the value of the acquisition function at that point, evaluated with the surrogate model.

Not only is the GA a natural choice for sequence optimization, it also permits easy incorporation of constraints. We can easily restrict the positions that we want to permit mutation in and restrict the allowed mutations in each location based on expert knowledge. Furthermore, we incorporate liability filtering to prevent the introduction of glycosylation sites and to exclude sequence motifs that are known to affect stability or other properties of the antibody.

The GA is modified for the batch BO version, where the acquisition function is jointly defined over a batch of sequences, i.e. $\alpha_q: \mathcal{S}^q \rightarrow \mathbb{R}$. Hence, the GA no longer evolves individual sequences, but batches of them. We introduce a batch-crossover operation that generates offspring batches from two parental batches by performing single-point crossover between sequences in the other batch and swapping sequences between batches with a batch crossover rate of $0.7$.

Instead of the acquisition function, we can directly interface the objective function as a fitness function in the GA. This makes the GA an obvious baseline to compare to. In multi-objective optimization, we employ a sum of normalized scores as the fitness function.

\subsection{Oracles} 
\label{sec:oracles}

\begin{description}
    \item[Affinity predictor] We train a neural network predictor on experimental affinity data for each considered antibody-antigen pair to predict relative improvement of affinity over the parental. To handle the small number of data points, we augment the dataset by considering the difference in affinity between sequence pairs, inspired by \citet{lin2025dyab}. Our model uses an AbLang-2 tokenizer \citet{olsen2024addressing} and a CNN-based regression head to predict the delta in binding affinity with respect to a reference sequence (e.g., parental sequence). 
    \item[Humanness score] We use \texttt{promb}'s implementation of the OASis score \citep{prihoda2022biophi}, a humanness score based on 9-mer peptide search in the Observed Antibody Space (OAS) \citep{kovaltsuk2018observed}.
    \item[Sequence likelihoods] We compute the mean log-probability of amino acids in sequences using the protein language model ESM-2 with 3B parameters \citep{lin2023evolutionary}.
    \item[Structure prediction] Structure prediction tools do not inherently predict antibody properties that can be measured in the laboratory. However, they provide scores that represent the confidence of the model about the predicted structure. We use the interface predicted TM score (ipTM) from Boltz-2 \citep{passaro2025boltz} to score model confidence at the interface between the antibody and the antigen, a metric with potential correlation to binding signal \citep{zambaldi2024novo}. With a runtime in the order of minutes per sequence on a GPU, this is the slowest objective we are considering.
\end{description}

Antibody lead optimization campaigns may target different sets of properties; one campaign might focus more on developability, while another might target cross-reactivity to multiple antigens, requiring a complete disparate set of \textit{in silico} predictors.

\method~provides a simple scoring function interface that makes interchanging scoring functions straightforward. Building purely on oracles, \method~does not directly address the issue of oracle quality - it is up to the user to decide which oracles to use for a particular campaign. This design choice reflects the fact that experimental measurements are costly and slow to obtain, and that \textit{in silico} predictors are imperfect but often serve as the best available proxies during lead optimization. We can leverage well-studied and sophisticated predictors instead of relying on predictors trained purely on small experimental datasets. In principle, the framework supports weighting oracles according to user preference - for example, the Expected Weighted Hypervolume Improvement criterion extends EHVI to accommodate weighted objectives \citep{feliot2018user}. 

\section{RELATED WORK}

\paragraph{Traditional and Evolutionary Baselines}\enlargethispage{\baselineskip} Genetic and evolutionary optimizers have long been used for optimizing black-box functions in discrete spaces; see \citet{katoch2021review} for a recent review. These algorithms gradually evolve a solution through random mutation, crossover and selection without gradients. Multi-objective extensions like NSGA-II \citep{deb2002fast} optimize for diverse Pareto-optimal solutions.

Traditional approaches are often inefficient in protein design given the high-dimensional sequence space \citep{turner2021bayesian}. This has motivated specialized protein evolutionary algorithms that incorporate domain-specific knowledge, such as AdaLead \citep{sinai2020adalead} and PEX \citep{ren2022proximal}. Some methods use neural networks or language models to guide mutation selection \citep{nigam2019augmenting, yang2019machine, nana2025integrating}. 

\paragraph{Generative Models and Reinforcement Learning (RL)}

Recent advances in generative modeling have enabled the synthesis of novel, plausible antibody sequences by learning from large corpora of protein data. Key approaches include transformer-based protein language models trained using either masked language modeling or next-token prediction objectives (in autoregressive models) \citep{rives2021biological, ferruz2022protgpt2, nijkamp2023progen2}. Most autoregressive language models are limited to generating full sequences, but there exist extensions for conditional infilling to redesign specific regions like CDRs \citep{shuai2023iglm, melnyk2023reprogramming}. Diffusion models represent another promising direction \citep{ho2020denoising}; see \citet{he2025ai} for a recent review. These models are increasingly fine-tuned to steer generation towards sequences or edits which respect developability constraints and optimize certain objectives \citep{goel2024token, yang2025steering}. RL offers a flexible framework to align pre-trained generative models with specific experimental objectives, often employing policy gradient methods such as REINFORCE to optimize the generation towards specific properties. Both proximal policy optimization (PPO) \citep{angermueller2019model, lee2025reinforcement} and Direct Preference Optimization (DPO) \citep{widatalla2024aligning} have been applied to single-objective protein design. 

\paragraph{Bayesian Optimization}\enlargethispage{\baselineskip} BO uses uncertainty-calibrated surrogates and acquisition functions for sample-efficient optimization of expensive black-box objectives. Rather than learning to generate sequences directly, BO methods iteratively propose candidates by balancing exploration of uncertain regions with exploitation of high-performing areas in the search space, and evaluate proposals using a predictive oracle or real experimental data. However, most current BO frameworks target single objectives or specific architectures, limiting applicability to multi-objective antibody design; \citep{gonzalez2024survey} provides a recent survey of BO methods for antibody design. A key challenge is the high-dimensional nature of discrete sequence space \citep{wang2016bayesian}.

Latent Space Bayesian Optimization employs a BO framework within a continuous latent space to search for optimal sequences. While many approaches use a Variational Autoencoder (VAE) pretrained on a large dataset \citep{gomez2018automatic, tripp2020sample, notin2021improving, lee2023advancing, moss2025return}, recent advances exploit the robust feature learning capabilities of Denoising Autoencoders (DAEs) \citep{maus2022local, stanton2022accelerating, gruver2023protein}. However, ensuring decoded sequences remain plausible is challenging. \citet{lee2025latent} approach this in single-objective sequence optimization using autoregressive normalizing flows to eliminate the reconstruction gap. These methods also require a very large training dataset.

Other approaches operate directly in sequence space using specialized kernels and discrete optimization methods. BOSS \citep{moss2020boss} employs string kernels with genetic algorithms for acquisition optimization, while AntBO \citep{khan2022antbo} uses a Transformed Overlap Kernel (TK) and a deep ProteinBERT \citep{brandes2022proteinbert} kernel for CDRH3 optimization. AntBO additionally employs trust-region-based search restrictions \citep{eriksson2019scalable}, which can help the search in high dimensions \citep{zhang2021unifying}. Recent work has also explored hybrid approaches that combine generative modeling with BO principles. CloneBO \citep{amin2024bayesian} combines language models trained on clonal families with Thompson sampling for biologically-informed optimization.

\paragraph{Multi-Objective Optimization}\enlargethispage{\baselineskip} 
Few approaches directly consider multi-objective optimization, where the hypervolume indicator is typically used to quantify the quality of the Pareto front. \citet{ren2025multi} incorporates multiple objectives as constraints in an RL framework, while other approaches use gradient-based optimizers requiring differentiable predictors or lengthy computation \citep{emami2023plug, luo2025pareto}. LaMBO and LaMBO-2 \citep {stanton2022accelerating, gruver2023protein} extend a BO framework with DAEs and generative infilling to multiple objectives, using expected hypervolume improvement (EHVI) as an acquisition function. ALLM-Ab \citep{furui2025allm} uses a fine-tuned protein language model in an active learning framework where sequences are selected based on hypervolume maximization. Our work represents the first multi-objective BO framework for optimizing black-box in-silico predictors directly in discrete sequence space.

\section{EXPERIMENTS}

We evaluate our multi-objective Bayesian optimization framework across three experimental settings: the optimization of a single-domain antibody with a focus on cross-reactivity, a benchmark comparison against LaMBO-2, and additional studies examining the impact of key design choices in the Appendix.

\subsection{Cross-reactivity of a V$_{\rm HH}$}
\label{sec:vhh}

We ran \method~on a therapeutic nanobody (V$_{\rm HH}$), a single-domain antibody derived from the heavy chain variable domain, to demonstrate the practical applicability of our framework to real-world antibody design scenarios. The lead optimization objective is to introduce cross-reactivity to two similar antigens, i.e., to enhance binding affinity on both while retaining or improving developability properties. We systematically optimize CDR1, CDR2, and CDR3 regions of the heavy chain individually, allowing up to 5 mutations per CDR region. The mutation space for each position was constrained to a curated dictionary of amino acids based on single-point mutations that we have experimental data for. This way we prevent reliance on an oracle that has not observed certain mutations, and because BOAT and all baselines use the same dictionary, it does not introduce bias into the comparative evaluation. This setting reduces the size of the search space and allows us to brute-force the computation of the complete `ground truth' Pareto front, defined as the Pareto front induced by exhaustively enumerated oracle scores, for up to 3 objectives.

Our goal is to evaluate whether \method~can efficiently recover the Pareto front as defined by the \textit{in silico} predictors and explore the sequence space to optimize the predictor values; the true experimental landscape is not available for these antibody systems. We emphasize that direct access to the `ground truth' Pareto front is rarely available, as the design space is typically vast and exhaustive evaluation using computational oracles is prohibitively expensive. We progressively increase the number of objectives from 2 to 4 to evaluate the scalability of \method~with problem dimensionality and compare sequential and batch design. For the main task of introducing cross-reactivity, we leverage two affinity predictors described in Section~\ref{sec:oracles} that were trained on the experimentally measured affinities for both antigens of 340 single-point and 26 quadruple mutations. We add the humanness score (third) and a PLM log-likelihood (fourth - where computing the ground truth was intractable) as additional objectives (cf. Section~\ref{sec:oracles}).

We benchmarked against two GA baselines. Our first GA baseline was set up as a standard GA (described in Section~\ref{sec:GA}) which optimizes a normalized sum of the objectives. Our second GA baseline is NSGA-II \citep{deb2002fast}, a GA specifically tailored for multi-objective optimization. NSGA-II maintains population diversity by mutating solutions along the Pareto frontier, though performance degrades with increasing objectives \citep{purshouse2003evolutionary}.

We evaluated our methods by comparing their discovered Pareto fronts to the ground truth where available, and track how the hypervolume evolves over oracle calls. By default, BOAT computes the reference point at the start of optimization as the minimum of the initially scored sequences minus $10\%$. For fair and consistent comparison of hypervolumes across different initializations, we fixed the hypervolume reference point across all experiments. We set it to $[-3,-3,0,-1]$ for the two affinity predictors, humanness score, and the PLM log-likelihood respectively, using prior knowledge of the oracle score ranges. The total number of `ground truth' sequences are 1,438,121, 33,829,027 and 61,602,147 for CDR1, CDR2 and CDR3 respectively. All methods have a budget of 1000 oracle calls. See Appendix~\ref{apx:crossreactivity} for further experimental details. We note that the ground truth hypervolume and search space is much higher for CDR3. It is easier to destabilise binding for CDR3, and CDR3 is known to be the most important for antigen recognition \citep{xu2000diversity}. We focus ablations on CDR3, comparing encodings and batch sizes for batch BO in Appendix~\ref{apx:encodings} and \ref{apx:batch_size}, respectively.

\subsubsection{Hypervolume evolution}

\begin{figure}[h!]
    \centerline{\includegraphics[width=0.95\columnwidth]{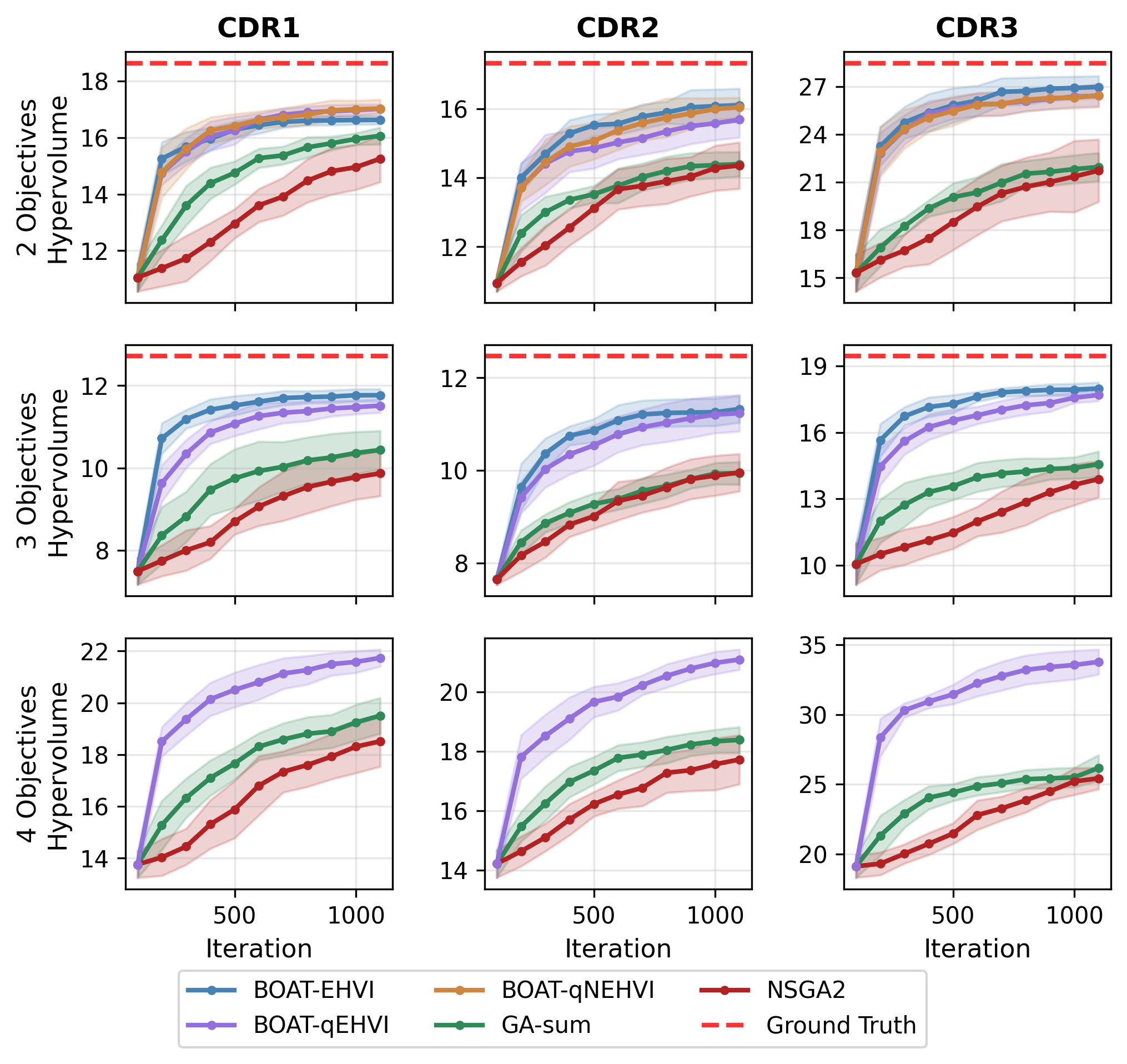}}
    \caption{Hypervolume vs. iterations (mean ± s.e., 10 seeds) for CDR1–3 under 2, 3, and 4 objectives. BOAT variants (q indicates batch acquisition functions) outperforms GA‑sum and NSGA‑II. Red dashed lines show ground‑truth hypervolume where available.} \label{fig:hypervolumecompare_allobj}
\end{figure}

Figure~\ref{fig:hypervolumecompare_allobj} shows the hypervolume evolution as a function of oracle calls for all methods across CDR1–CDR3 and across 2, 3, and 4 objectives. BOAT variants consistently reach higher hypervolume earlier and achieve larger final hypervolume than GA baselines, and maintains its effectiveness even in higher-dimensional objective spaces. Batch acquisition (qEHVI, qNEHVI) favours broader exploration early with a lower hypervolume, but there was no significant difference in the final hypervolume found between acquisition functions. NSGA‑II underperforms progressively as the number of objectives increases, consistent with prior reports.

qNEHVI became much slower when increasing the number of objectives; while two-objective qNEHVI evaluations complete in seconds, three objectives requires several minutes per BO step ($>$ 100 times slower). This rapid degradation is consistent with findings in \citet{daulton2021parallel}. As qNEHVI did not outperform qEHVI in the 2-objective setting, we did not use this for more objectives. For the 4‑objective setting, querying the PLM is the bottleneck, so we report batch BOAT versus GA baselines only.

\subsubsection{Validation against ground truth, sequence diversity, and PLM}

\begin{figure}[h!]
    \centering
    \begin{subfigure}[b]{\columnwidth}
        \centering
        \includegraphics[width=0.85\columnwidth]{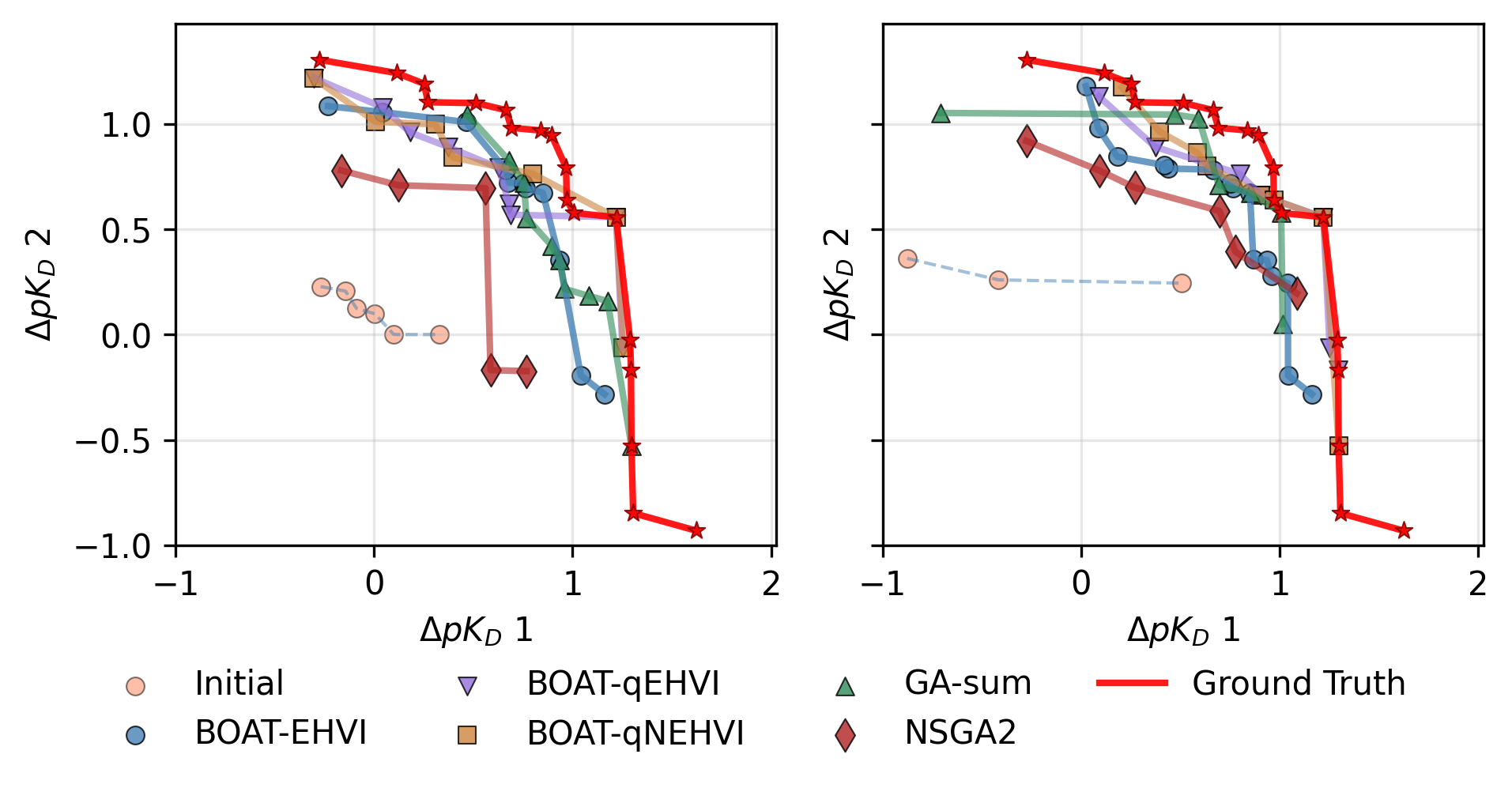}
        \vspace{-1ex}
        \caption{CDR1}
        \label{fig:paretocdr1}
    \end{subfigure}\vspace{1ex}
    \begin{subfigure}[b]{\columnwidth}
        \centering
        \includegraphics[width=0.85\columnwidth]{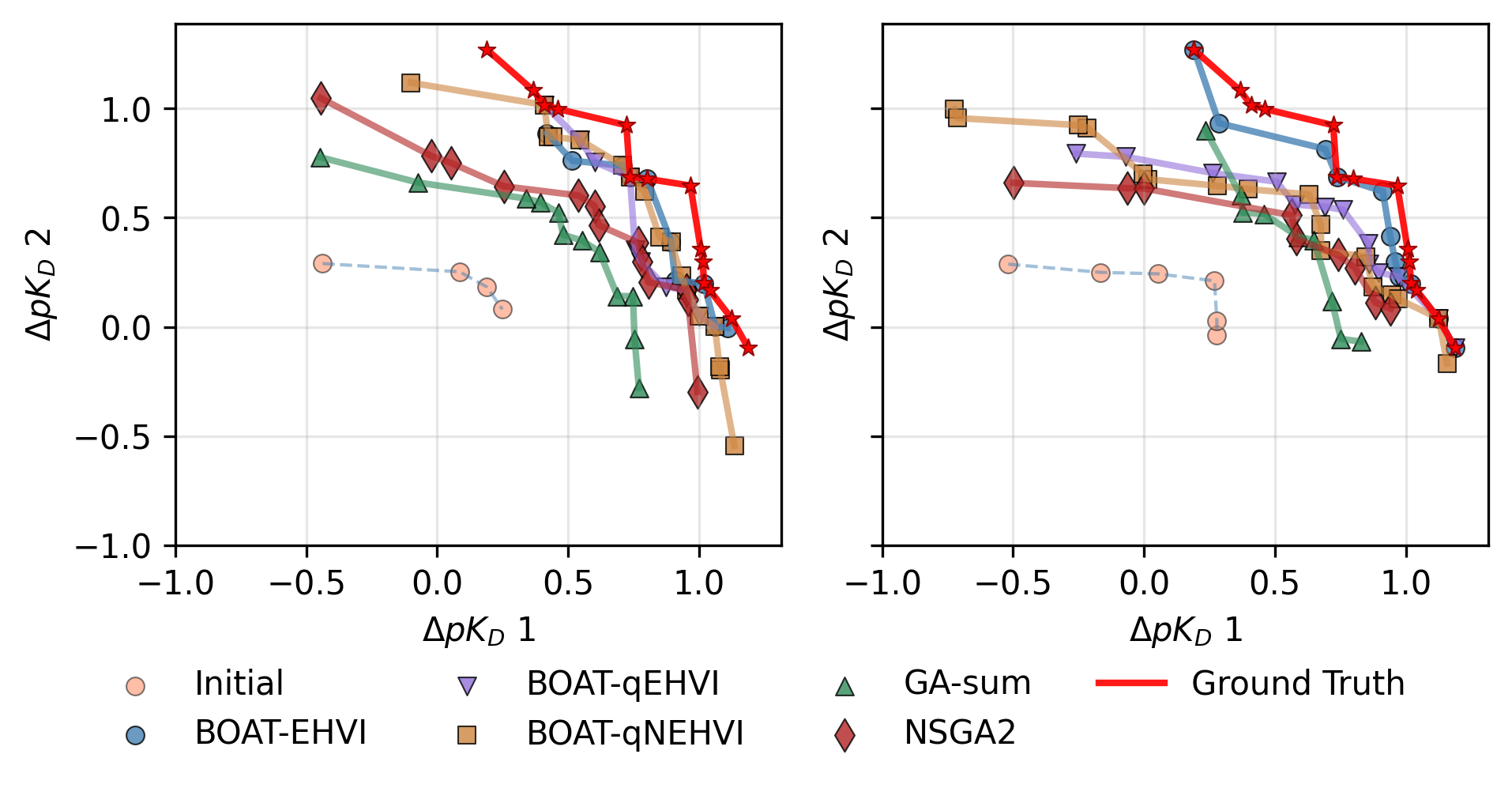}
        \vspace{-1ex}
        \caption{CDR2}
        \label{fig:paretocdr2}
    \end{subfigure}\vspace{1ex}
    \begin{subfigure}[b]{\columnwidth}
        \centering
        \includegraphics[width=0.85\columnwidth]{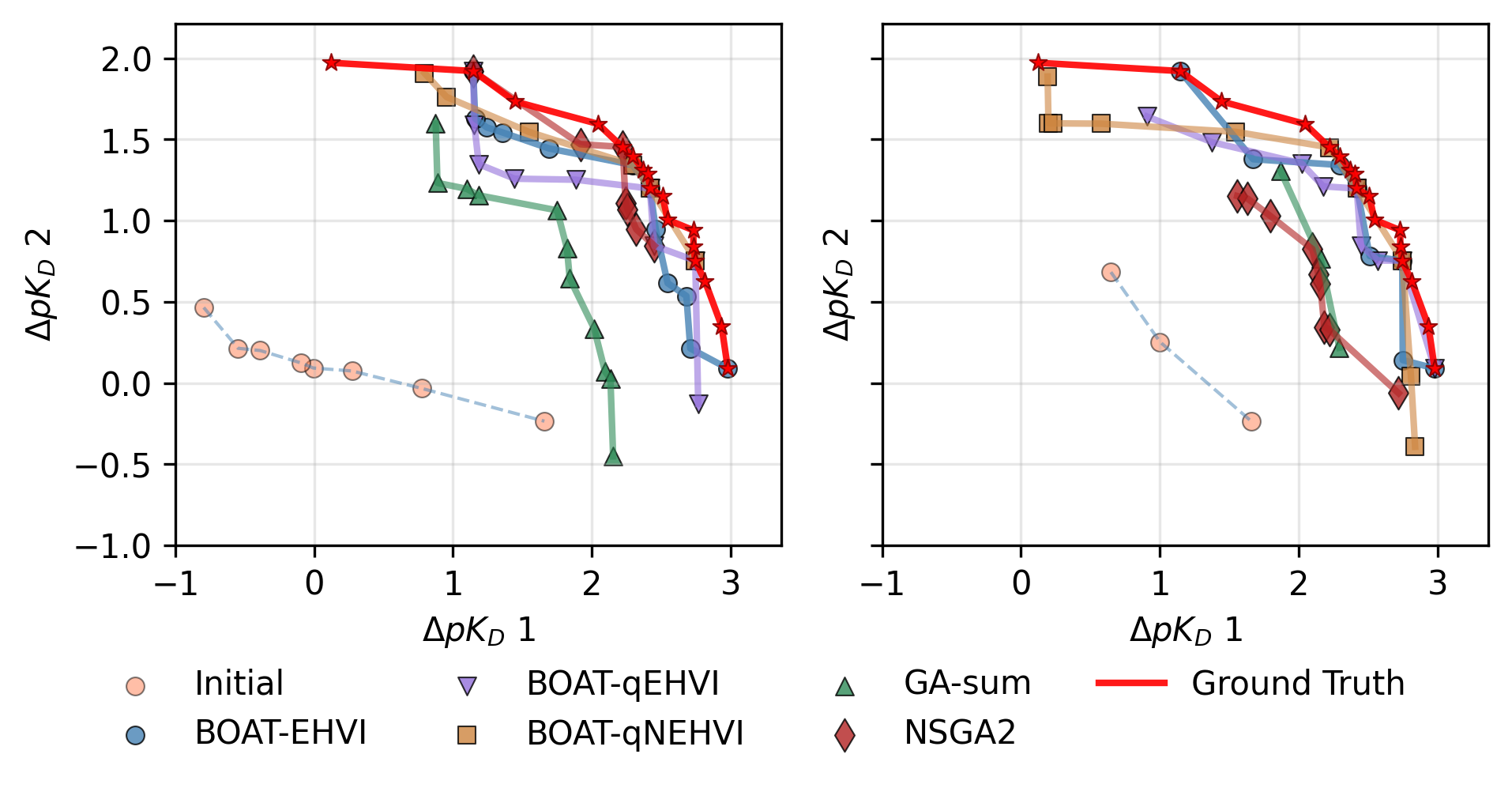}
        \vspace{-1ex}
        \caption{CDR3}
        \label{fig:paretocdr3}
    \end{subfigure}
  
    \caption{Discovered vs. ground truth Pareto fronts for 2-objective CDR optimization with 5 mutations. Each panel shows the best-performing seed for GA methods (left) and \method~(right). We can see that even the most successful GA runs fall behind \method.}
    \label{fig:5mut}
\end{figure}

In 2 dimensions, we can visualize the discovered Pareto front against the `ground truth' Pareto front derived from exhaustive evaluation. A subset of these plots is visualized in Figure~\ref{fig:5mut}, displaying the seed with the highest hypervolume among all GA methods and the seed with the highest hypervolume among all \method~variants for that CDR. Plots for all seeds and also plots for earlier points in all the models are in Appendix~\ref{apx:allseeds}. \method~traces fronts close to the true frontier and often recovers true Pareto‑optimal sequences even in CDR3’s 63M‑sequence space.

\begin{figure}[h!]
    \centerline{\includegraphics[width=\columnwidth]{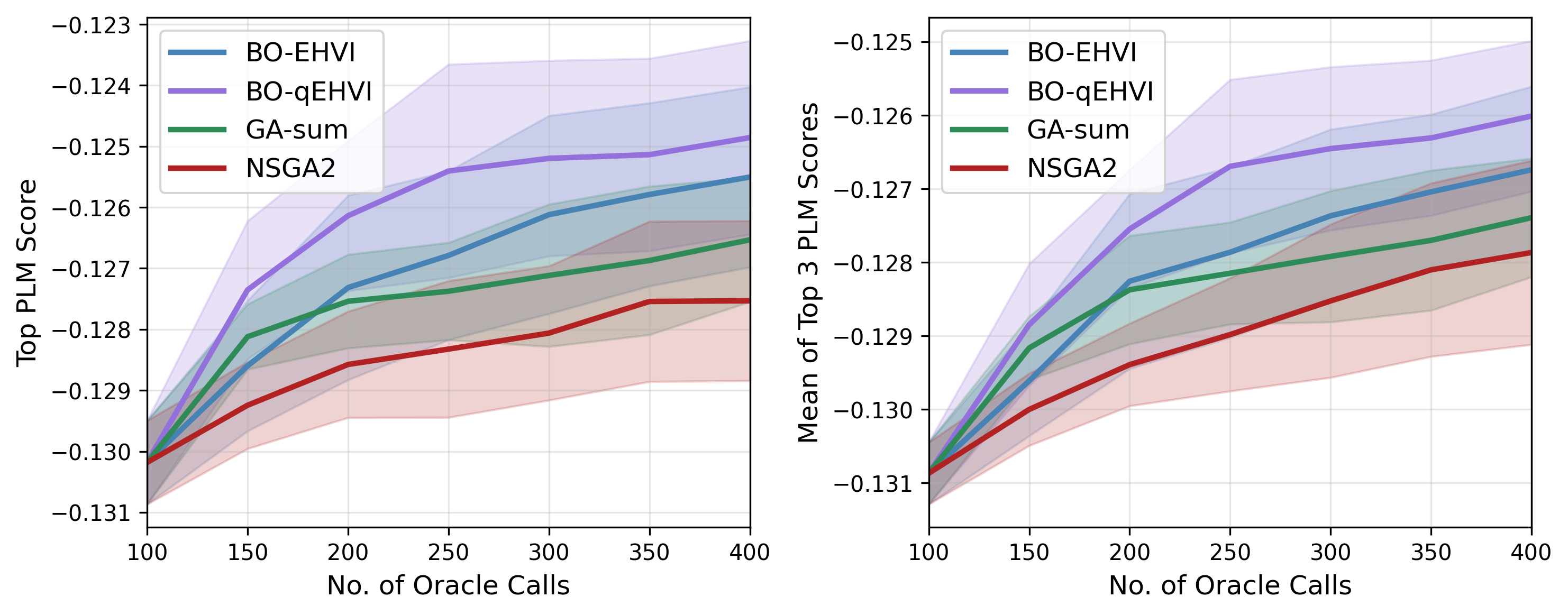}}
    \caption{PLM evolution for the first 300 generations of 3-objective CDR3 optimization, with the best PLM score of all generated sequences and the mean score of the top 3 PLM scores recorded. Results averaged over 10 seeds with standard error. Note that the x-axis starts at 100 to omit the initial sequences.} \label{fig:3obj_plm}
\end{figure}

To evaluate whether our multi-objective optimization approach leads to improved fitness beyond the explicitly optimized objectives, we scored the first 300 generated sequences from each method using ESM-2 in the 3-objective setting. Figure~\ref{fig:3obj_plm} reveals that \method~produces sequences with slightly higher PLM scores early in the optimization process without explicitly optimizing for this. Multi-objective BO naturally favours sequences with better biological fitness, even when all methods operate under identical mutation constraints.

\begin{figure}[h!]
    \centering
    \begin{subfigure}[b]{0.45\columnwidth}
        \centering
        \includegraphics[width=\columnwidth]{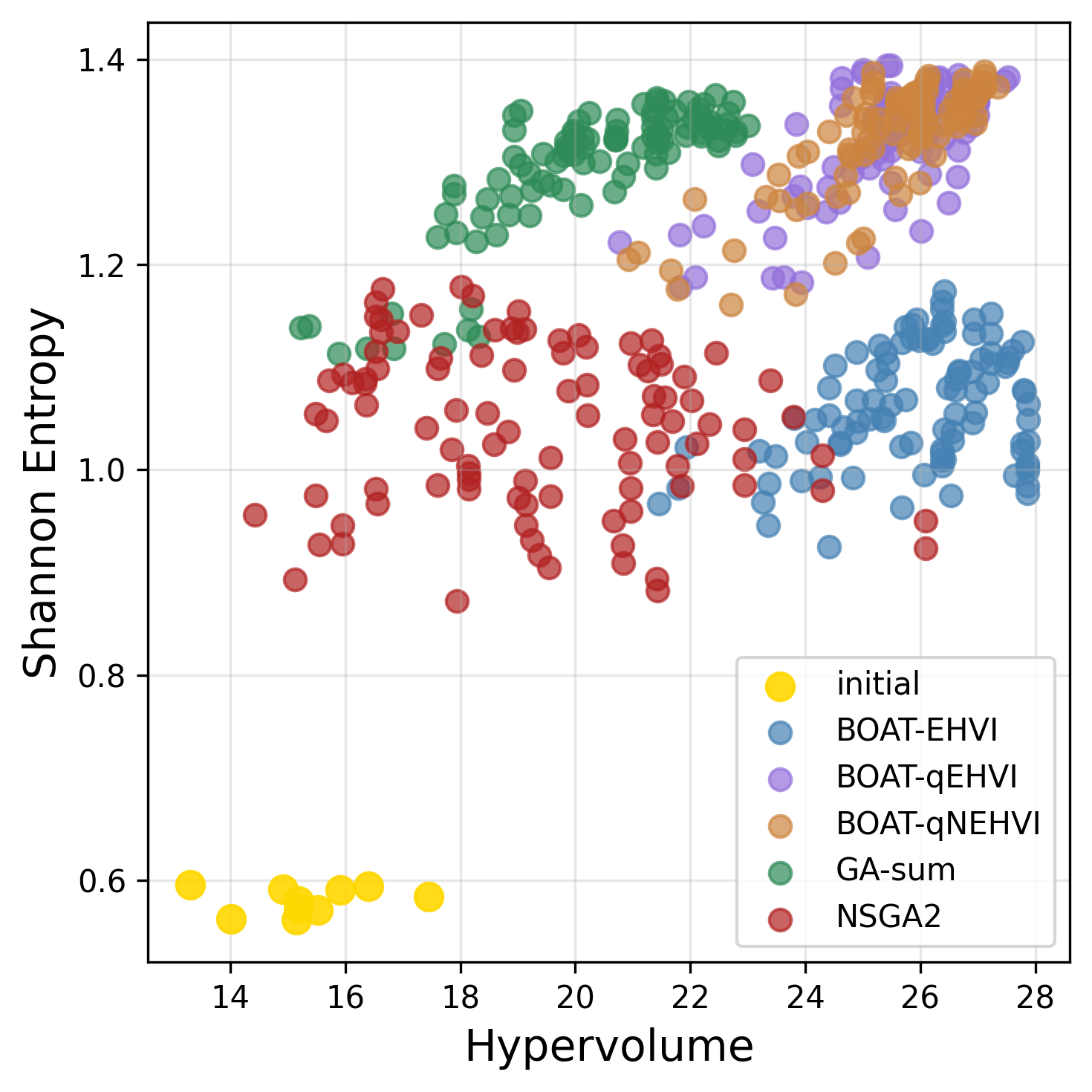}
        \caption{Diversity vs. HV}
        \label{fig:shannonentropy}
    \end{subfigure}
    \hfill
    \begin{subfigure}[b]{0.45\columnwidth}
        \centering
        \includegraphics[width=\columnwidth]{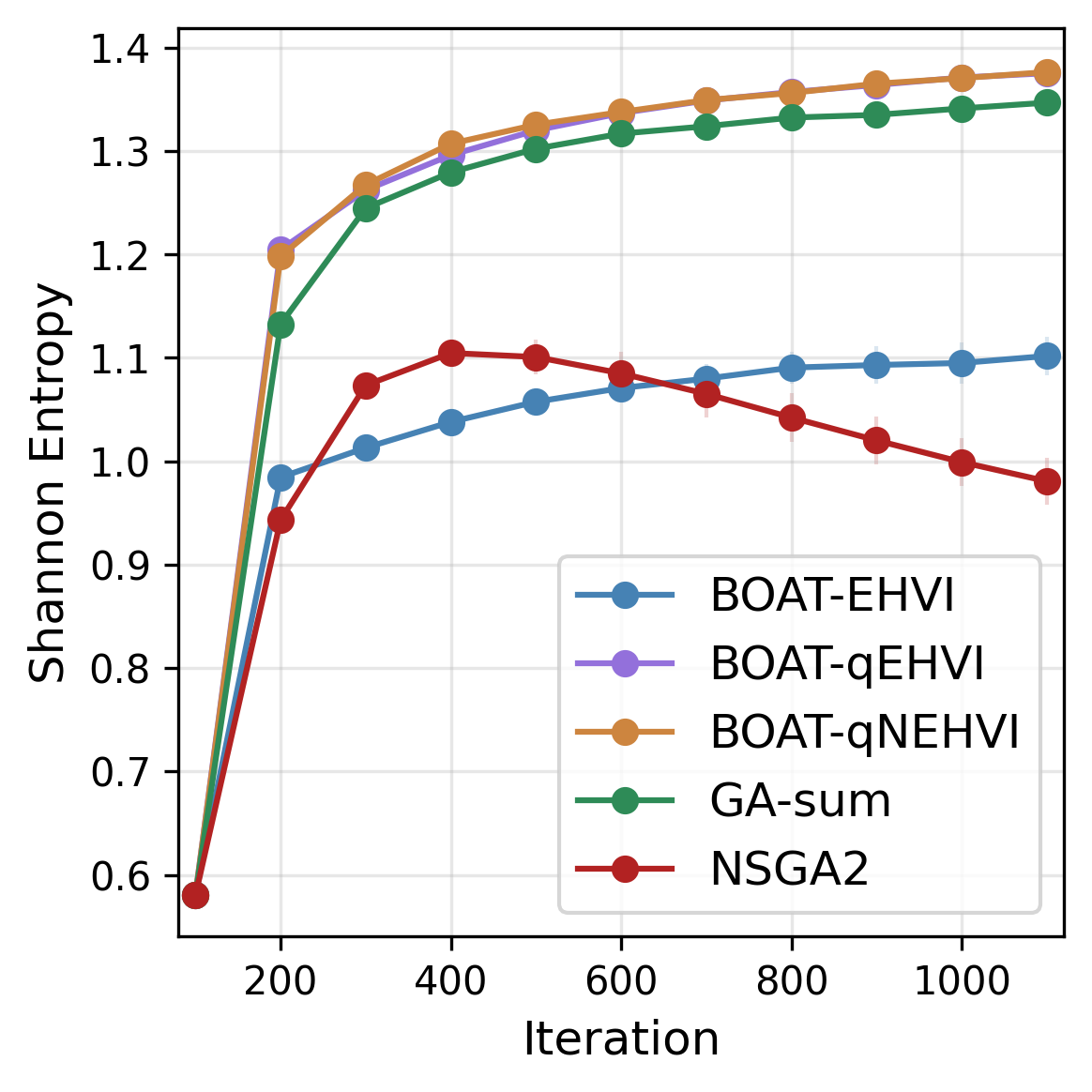}
        \caption{Evolution of diversity}
        \label{fig:shannonvsiter}
    \end{subfigure}
    \caption{(a) Scatter plot of hypervolume versus Shannon entropy for all seeds, methods, and every 100 iterations for CDR3. Initial solutions are highlighted in gold. Each point represents the diversity and multi-objective performance of a population at a given optimization step. (b) Evolution of Shannon entropy over optimization iterations for CDR3. Results averaged over 10 seeds with standard error bands.}
    \label{fig:shannonentropyfull}
\end{figure}

To assess the diversity of solutions discovered by each method, we computed the average Shannon entropy for all generated sequences for every 100 generated sequences, for all methods in the 2-objective setting. A visualization of the results for CDR3 can be seen in Figure~\ref{fig:shannonentropy}, comparing Shannon entropy to hypervolume and iteration, with additional figures for other CDRs in Appendix~\ref{apx:shannon}. Batch acquisition methods (\method-qEHVI, \method-qNEHVI) achieve the optimal combination of both high hypervolume performance and high sequence diversity. The larger diversity likely stems from the batch acquisition's inherent mechanism of selecting multiple diverse candidates simultaneously, naturally promoting exploration of different regions of the sequence space. \method-EHVI, while achieving competitive hypervolume performance, exhibits lower sequence diversity, suggesting more focused exploitation around promising regions. While both GA methods are inferior in hypervolume performance, it is interesting that GA-sum is able to explore more diverse sequences compared to NSGA-II. High sequence diversity is crucial for experimental validation campaigns, as it provides multiple distinct candidates for testing while maintaining optimization quality. Figure~\ref{fig:shannonvsiter} shows that \method~successfully maintains sequence diversity throughout the algorithm.

\subsubsection{Structure prediction oracle} \label{sec:boltz}
\begin{figure}[h!]
    \centerline{\includegraphics[width=0.7\columnwidth]{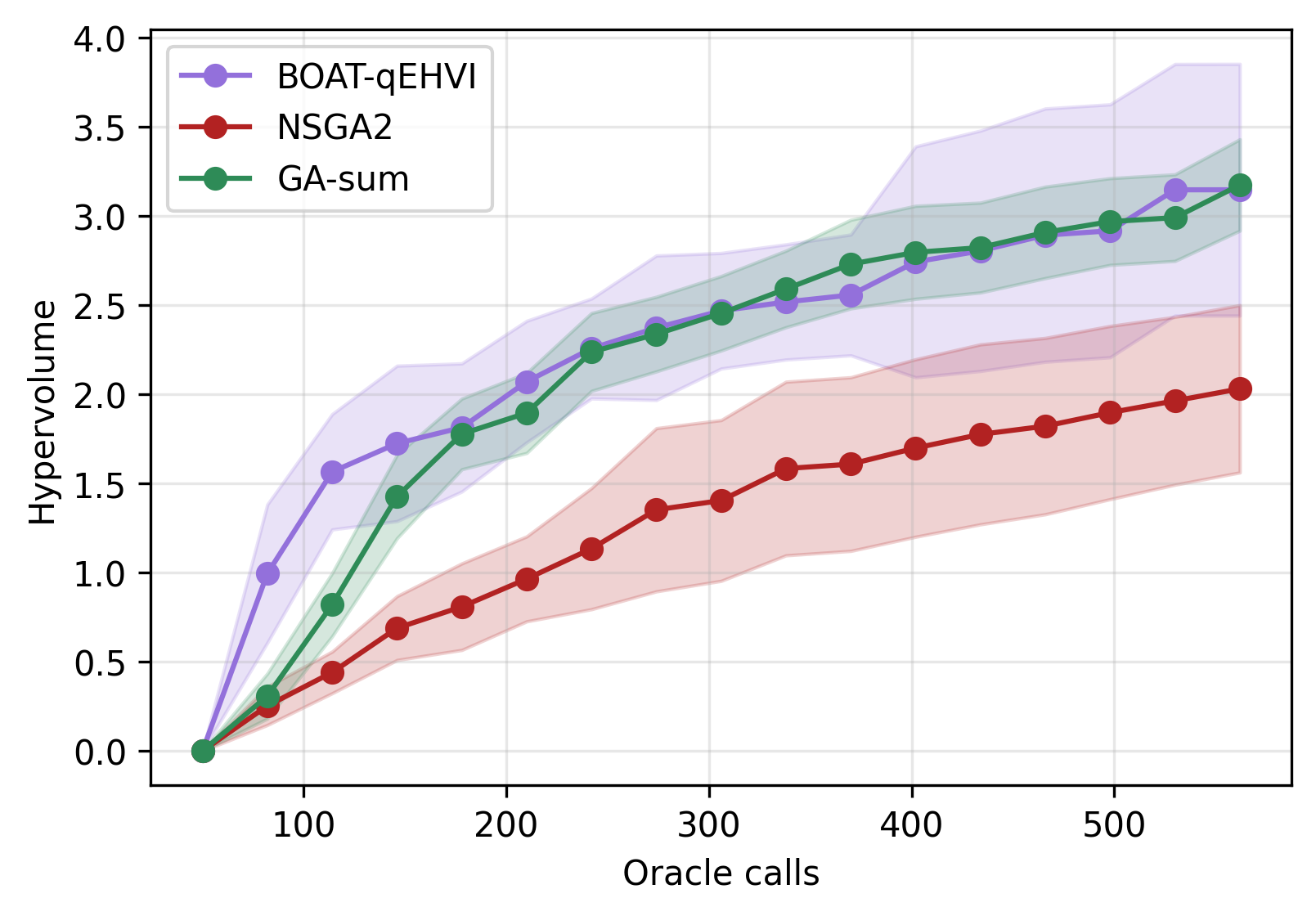}}
    \caption{Hypervolume evolution for 3-objective CDR optimization. Initial HV has been subtracted. Results averaged over 5 seeds with standard error.} \label{fig:boltz_hypervolume}
\end{figure}

We further consider a 3-objective setup with two affinity predictors and Boltz-2 for structure prediction, where we considered ipTM the score of interest to optimize; further details are in Appendix~\ref{apx:boltz}. Boltz-2 metrics are challenging oracles that rely on a powerful 3D antibody-antigen representation from the AlphaFold Pairformer. Unsurprisingly, the Tanimoto model with BLOSUM encoding struggles to capture this structural complexity. The GA interestingly achieves comparable performance to Bayesian optimization methods via semi-random mutations in this scenario, cf.~Figure~\ref{fig:boltz_hypervolume}. Remarkably, NSGA-II does consistently worse here.

\subsection{4-4-20 scFv Antibody}
\label{sec:lamboexperiment}

\subsubsection{Dataset and Experimental Setup}
We now run a comparison of multi-objective optimization between \method~and LaMBO-2 \citep{gruver2023protein}. LaMBO-2 requires a corpus of training data to be used as it trains its own predictive model $f^\ast$, rather than allowing for the use of any external oracle in its optimization loop. We use a public dataset from \citet{adams2016measuring}, which is featured in the Fitness Landscape of Antibodies (FLAb) benchmark \citep{chungyoun2024flab}. This consists of 10K+ affinity and expression measurements derived from mutating CDR1 and CDR3 regions of the 4-4-20 scFv antibody \citep{boder1997yeast}. This dataset contains between 3 and 18 repeated measurements for each antibody, so we take a mean of the repeats and retain only sequences with both affinity and expression measurements, resulting in 2807 total sequences. We extracted the predictive model $f^\ast$ from LaMBO-2's trained discriminative head as the black-box oracles used in \method~for prediction of affinity and expression, enabling a fair comparison between the methods.

We generated 256 sequences from each method; further details can be found in Appendix~\ref{apx:lambo_exp_details}. We limit mutations to CDR1 and CDR3 and allow BOAT to introduce up to 8 mutations, a constraint used in a lead optimization study in \citep{gruver2023protein}. This allows for more mutational freedom than previously as we edit two CDRs. 

\subsubsection{Results}

\begin{wrapfigure}{r}{0.48\columnwidth}
    \vspace{-8pt}  
    \includegraphics[width=0.46\columnwidth]{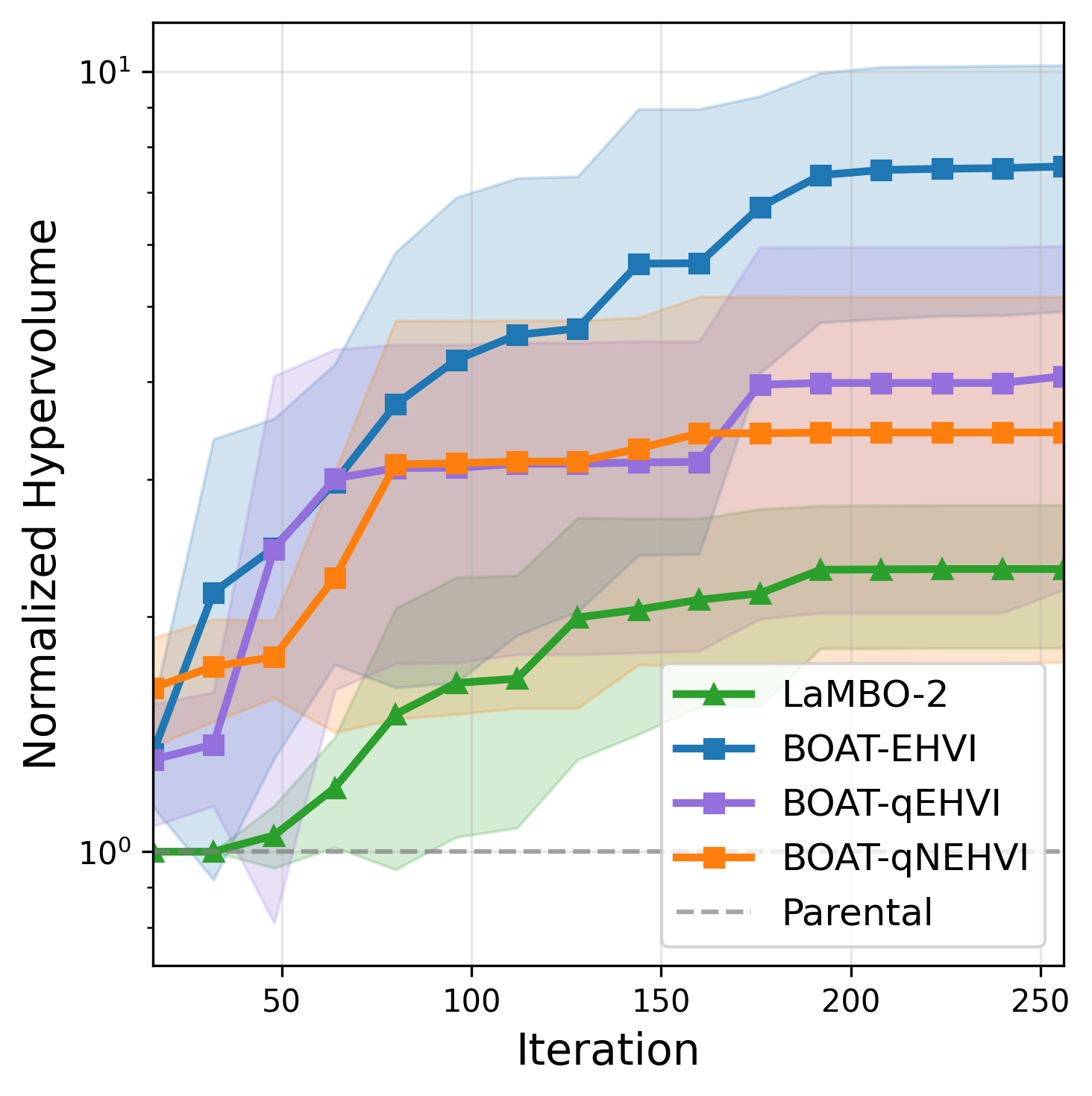}
    \vspace{-5pt}   
    \caption{Comparing relative hypervolume versus number of generated sequences between \method~and LaMBO-2. The reference hypervolume of 1 has been assigned to the parental sequence.}
    \label{fig:lambo_hypervolume}
    \vspace{-8pt}  
\end{wrapfigure}

We see from Figure~\ref{fig:lambo_hypervolume} that the hypervolume found by \method~is generally larger than for LaMBO-2 throughout optimization. The predicted scores of the generated sequences indicate that \method~is often able to push further toward Pareto-optimality than LaMBO-2. However, LaMBO-2's saliency-guided editing mechanism encourages mutations to amino acids observed in the training data, potentially generating more biologically realistic sequences backed with training data, whereas \method~does not impose biological priors. This may enable exploration of less common, yet beneficial mutations. 

To constrain BOAT to more realistic sequences, we added ESM-2 likelihood as a third oracle, but only compare the hypervolume given by the affinity and expression predictions in Figure~\ref{fig:lambo_hypervolume}. This counteracts the tendency of BOAT to discover sequences with implausible predictive values relative to the training distribution. These artifacts arise due to overfitting and mislead the optimization process toward sequences with artificially inflated predicted performance. Given the limited and imbalanced training data, such behaviour is not unexpected but reflects common issues in \textit{in silico} lead optimization. Appendix~\ref{apx:lambo_results} contains further insights and discussion about the respective merits of LaMBO-2 and \method.

This comparison has inherent limitations that merit discussion. LaMBO-2's architecture tightly couples a generative diffusion model with a discriminative head through shared layers, making it impossible to substitute external oracles for evaluation. LaMBO-2 is elegant in that it condenses the entire computational antibody design into a single procedure without human intervention. This is powerful for designing sequences purely from experimental measurements of candidate traits. However, the human-in-the-loop design approach taken with \method~is preferential to leverage both expert knowledge and state-of-the-art external predictors for diverse properties, some of which may not be directly measured in wet-lab experiments. Our plug-and-play approach addresses this limitation by treating predictors as modular components that can be easily swapped or added, while achieving comparable or superior performance in identifying high-performing sequences.

\section{DISCUSSION}

In this work, we presented \method, a lightweight plug-and-play multi-objective Bayesian optimization framework for antibody lead optimization that enables efficient exploration of sequence space to optimize Pareto trade-offs between multiple objectives. \method~allows users to interface arbitrary tools for antibody property prediction, enabling the joint optimization of existing state-of-the-art \textit{in silico} oracles.
\method~operates directly in sequence space, does not require any pre-training, and performs competitively with only a few initial scored sequences.
The success on two antibody candidates highlights a critical gap in current protein design methodology: most existing approaches optimize single objectives or require extensive pre-training, yet real-world antibody development demands simultaneous optimization of multiple properties with small experimental budget and long timescales. \method~bridges this gap through its modular design, allowing users to easily interface external oracles - which are not limited to those included in this work. This flexibility makes it practical for real-world antibody design.

While being simple and versatile, \method~has important limitations and potential for future research.
We observed in Section~\ref{sec:boltz} that the oracle's complexity can compromise the performance of the \method's surrogate model. A step ahead would be to leverage more tailored surrogates for antibody modelling, either through protein-specific kernels \citep{groth2024kermut} or encodings that capture antibody structure \citep{malherbe2024igblend}.
Section~\ref{sec:lamboexperiment} revealed that biological priors can prevent the generation of out-of-distribution sequences, and could be straightforwardly included in the GA by sampling from a PLM likelihood instead of uniformly.
Other areas of research can look into other acquisition functions; trust region Bayesian optimization \citep{eriksson2019scalable} has seen success in other high-dimensional regimes. While \method~still explored sequence space well with 4 objectives, it is known that GPs perform poorly in very high dimensions \citep{binois2022survey}. While it is likely that some filtering by certain properties will take place in real-world protein design if more properties were desired, we could explore further extensions to the Tanimoto GP used.

As a predominant challenge inherent to \emph{in silico} lead optimization, \method~accepts oracle predictions as the `ground truth'. The Pareto front found for computational oracles might only poorly represent the true underlying experimental Pareto front if the predictive power of the oracles is poor. Yet, the discrepancy between \textit{in silico} predictions and experimental measurements, especially for affinity, is a common issue in antibody design, aggravated by data scarcity. Hence, the ultimate experimental performance of \method~hinges on oracle quality. An exciting yet challenging direction of future research might be to inform \method~surrogates with experimental data and leverage the learned correlation with oracles to directly optimize experimental properties. Besides data availability, a strong inductive bias will be needed to leverage correlations between experiments and \textit{in silico} predictors successfully.

\subsection*{Acknowledgements}
JR acknowledges the support of this work through an internship at AstraZeneca, Cambridge, UK.
The authors thank Dino Ogli\'c, Owen Vickery, and Isabelle Sermadiras for valuable discussions, as well as Tom Diethe for constructive feedback on the manuscript.

\bibliography{boat}

\include{content/checklist}

\clearpage

\appendix
\include{content/supplement}

\end{document}

%% file: content/checklist.tex
\section*{Checklist}

\begin{enumerate}

  \item For all models and algorithms presented, check if you include:
  \begin{enumerate}
    \item A clear description of the mathematical setting, assumptions, algorithm, and/or model. \textbf{Yes}
    \item An analysis of the properties and complexity (time, space, sample size) of any algorithm. \textbf{No - time is dominated by the oracle of interest, and complexity of GPs is known.}
    \item (Optional) Anonymized source code, with specification of all dependencies, including external libraries. \textbf{Will be made available upon acceptance.}
  \end{enumerate}

  \item For any theoretical claim, check if you include:
  \begin{enumerate}
    \item Statements of the full set of assumptions of all theoretical results. \textbf{Not Applicable}
    \item Complete proofs of all theoretical results. \textbf{Not Applicable}
    \item Clear explanations of any assumptions. \textbf{Not Applicable}
  \end{enumerate}

  \item For all figures and tables that present empirical results, check if you include:
  \begin{enumerate}
    \item The code, data, and instructions needed to reproduce the main experimental results (either in the supplemental material or as a URL). \textbf{Instructions are available. Code will be made available upon acceptance.}
    \item All the training details (e.g., data splits, hyperparameters, how they were chosen). \textbf{Yes - throughout Section 4 and in Appendix 1, 7}
    \item A clear definition of the specific measure or statistics and error bars (e.g., with respect to the random seed after running experiments multiple times). \textbf{Yes - throughout Section 4}
    \item A description of the computing infrastructure used. (e.g., type of GPUs, internal cluster, or cloud provider). \textbf{No. The main algorithm can be run on a local CPU in seconds, resource requirements depend on oracles used.}
  \end{enumerate}

  \item If you are using existing assets (e.g., code, data, models) or curating/releasing new assets, check if you include:
  \begin{enumerate}
    \item Citations of the creator If your work uses existing assets. \textbf{Yes}
    \item The license information of the assets, if applicable. \textbf{No. All external packages used are cited and open source.}
    \item New assets either in the supplemental material or as a URL, if applicable. \textbf{No - Code and license will be made available upon acceptance.}
    \item Information about consent from data providers/curators. \textbf{Not Applicable}
    \item Discussion of sensible content if applicable, e.g., personally identifiable information or offensive content. \textbf{Not Applicable}
  \end{enumerate}

  \item If you used crowdsourcing or conducted research with human subjects, check if you include:
  \begin{enumerate}
   \item The full text of instructions given to participants and screenshots. \textbf{Not Applicable}
   \item Descriptions of potential participant risks, with links to Institutional Review Board (IRB) approvals if applicable. \textbf{Not Applicable}
    \item The estimated hourly wage paid to participants and the total amount spent on participant compensation. \textbf{Not Applicable}
  \end{enumerate}

\end{enumerate}

%% file: content/supplement.tex
%
\runningtitle{BOAT: Antibody Design via Multi-Objective Bayesian Optimization}

%

\onecolumn
\aistatstitle{BOAT: Navigating the Sea of In Silico Predictors for Antibody Design via Multi-Objective Bayesian Optimization: \\ Supplementary Material}

\section{FURTHER EXPERIMENTAL DETAILS - CROSS-REACTIVITY}
\label{apx:crossreactivity}

\subsection{5 maximum mutations}

In all experiments, we generate 100 initial sequences with 2 maximum mutations for each of 10 different random seeds per method. All methods were allowed up to 1000 oracle calls to evaluate sequences. Batch acquisition functions (indiciated by a `q' in the acquisition function; qEHVI and qNEHVI) had a batch size of 4, so were run for 250 iterations; sequential EHVI was run for 1000 iterations. All GAs (baselines and within \method) scored 50 sequences per generation over 20 generations. We used one-hot encoding. GA settings were as in Section~\ref{sec:vhh} for both the GA baseline and within the Bayesian optimization loop, except that the mutation probability was set as 0.15 for all GAs except BOAT runs with the qEHVI acquisition. This was to promote diversity due to the large number of iterations that other algorithms were run for. For NSGA-II, we use the version implemented in PyMoo \citep{blank2020pymoo} with custom mutation and crossover functions appropriate for sequences.

The GA comparison in this experimental setup is feasible as the objectives used in this section are fast to evaluate. Running the GA within the inner-loop of the Bayesian optimization takes less than one second in most cases. However, we reiterate that GAs are generally not suitable for tasks when objective functions require expensive experimental evaluation or lengthy simulations, highlighting a key advantage of BO.

\subsection{Structure prediction oracle}
\label{apx:boltz}

We considered a total of 512 oracle calls after the initial 50 evaluations. This translates to 16 rounds of 32 new sequences in the GAs, and 64 iterations of \method~using a batch size of 8. The GA for optimizing the acquisition function had a budget of $50 \times 50$ sequences in every \method~iteration.

\section{SHANNON ENTROPY FOR OTHER CDRs}
\label{apx:shannon}

\begin{figure}[ht]
    \centering
    \begin{subfigure}{0.3\columnwidth}
        \centering
        \includegraphics[width=\linewidth]{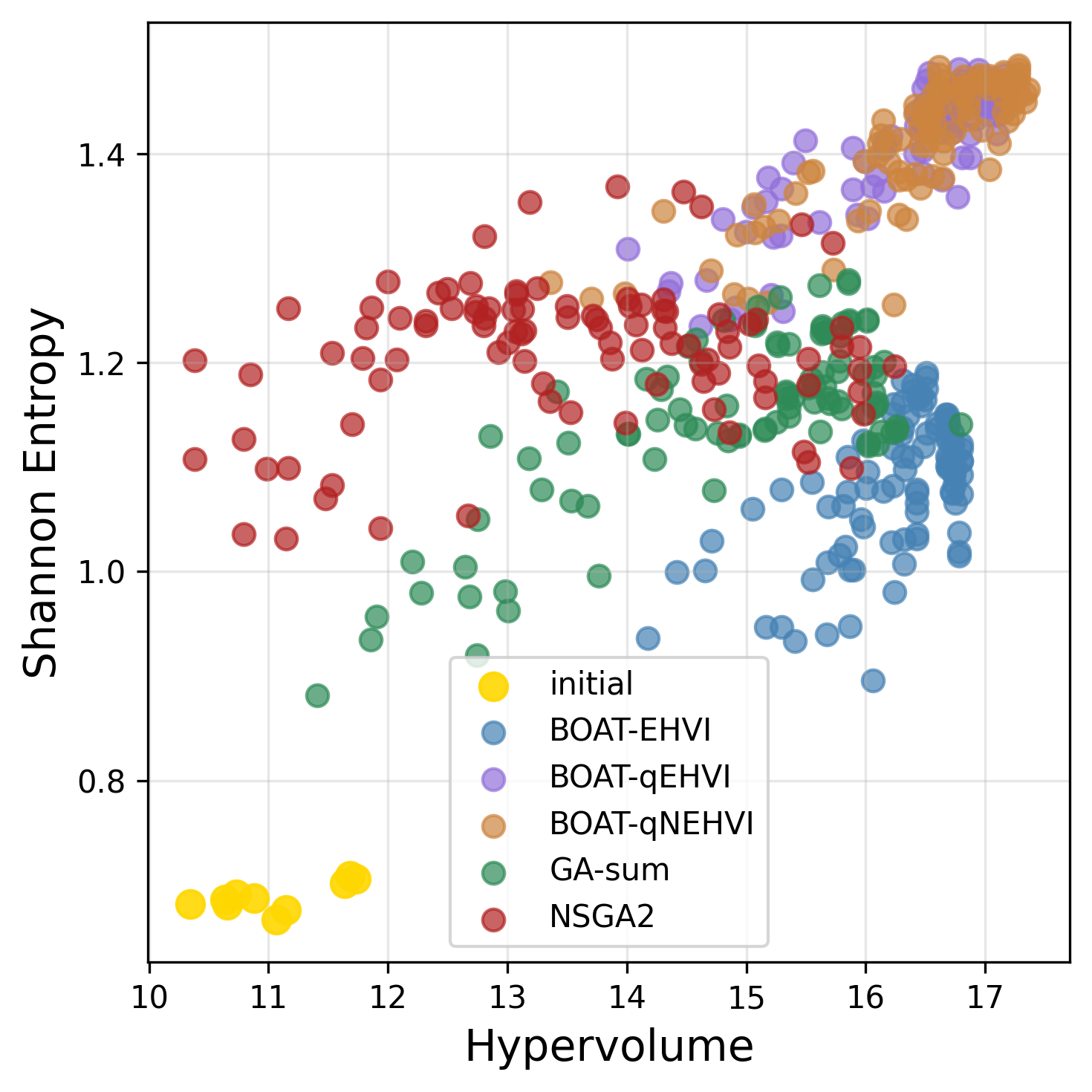}
        \caption{CDR1}
        \label{fig:entropy_cdr1}
    \end{subfigure}
    \hspace{0.05\columnwidth}
    \begin{subfigure}{0.3\columnwidth}
        \centering
        \includegraphics[width=\linewidth]{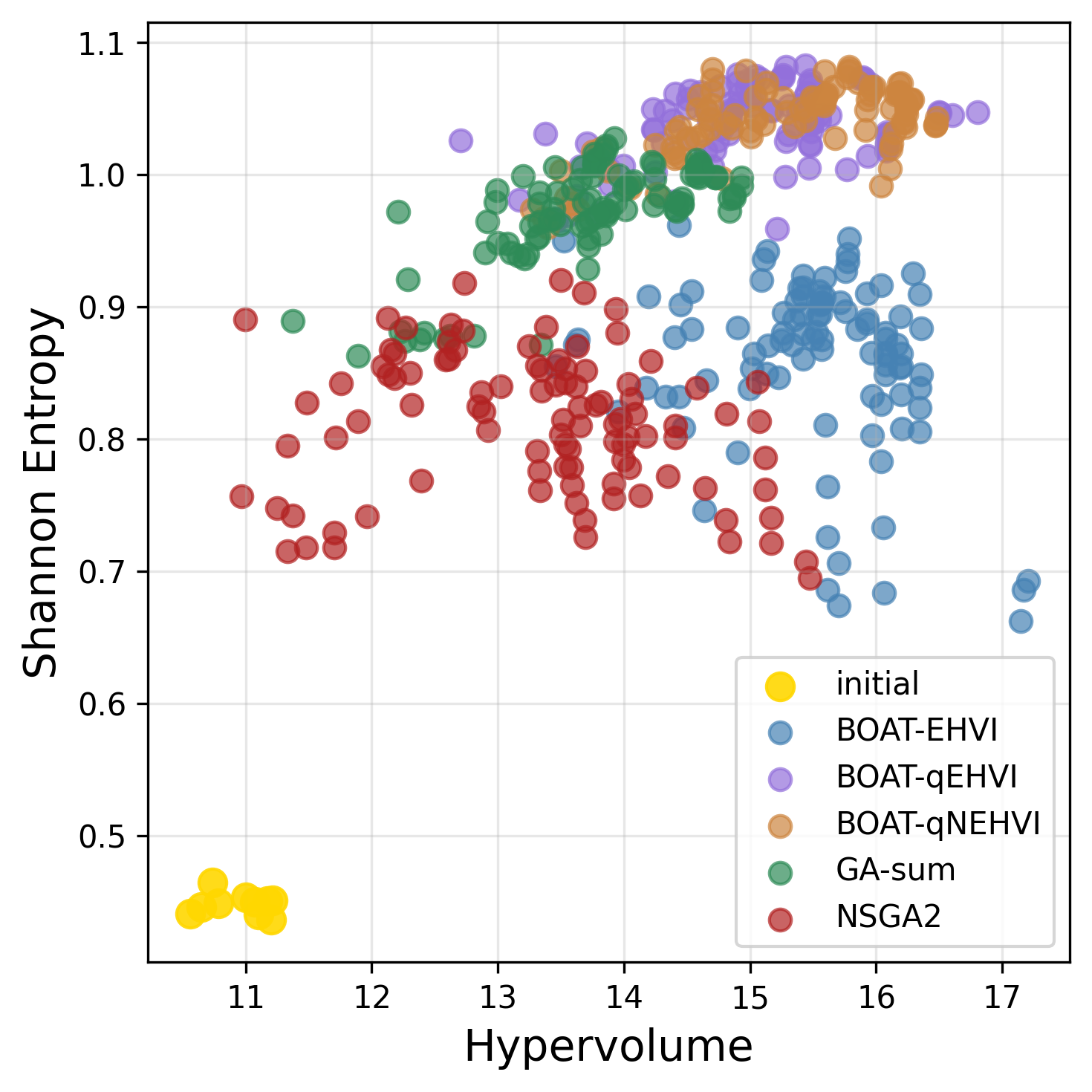}
        \caption{CDR2}
        \label{fig:entropy_cdr2}
    \end{subfigure}
    \caption{Scatter plots of hypervolume versus Shannon entropy for CDR1 and CDR2 optimization. Each point represents the diversity and multi-objective performance of a population at a given optimization step, showing results for all seeds, methods, and every 100 iterations. Initial solutions are highlighted in gold.}
    \label{fig:entropy_cdr1cdr2}
\end{figure}

\begin{figure}[h!]
    \centerline{\includegraphics[width=\columnwidth]{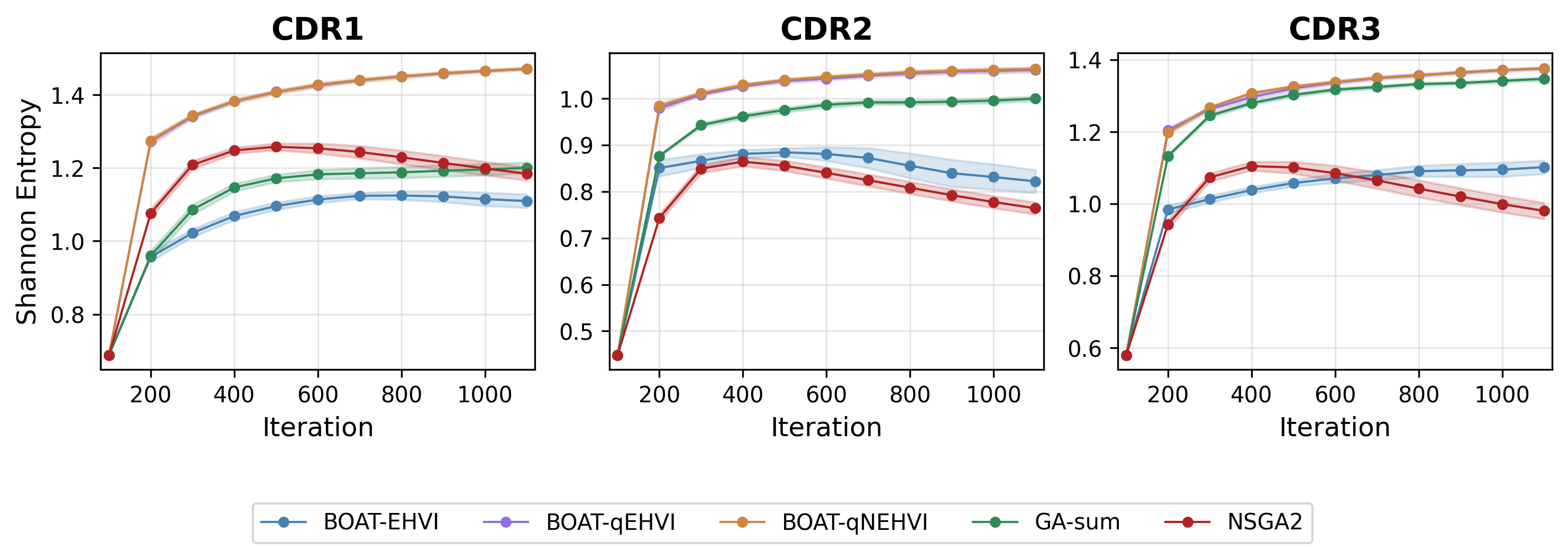}}
    \caption{Evolution of Shannon entropy over optimization iterations for CDR1, CDR2, and CDR3 regions. Results averaged over 10 seeds with standard error bands, showing how population diversity changes throughout the optimization process across all three CDR regions. Note BOAT-qEHVI performs almost exactly in line with BOAT-qNEHVI.} \label{fig:entropycombinediter}
\end{figure}

\section{COMPARING ENCODINGS}
\label{apx:encodings}

We compared the performance of different encodings (see Section~\ref{sec:vhh}) across 10 seeds when optimizing CDR3 for the V$_{\rm HH}$ antibody. We allowed for 5 maximum mutations on CDR3, and optimized for 2 affinity objectives. For each encoding, we tested EHVI, qEHVI and qNEHVI (with batch sizes 4), apart from AbLang-2, where we only tested qEHVI, as the computational cost of querying a language model made it significantly slower than all other encodings; each run of a GA took up to two minutes compared to less than five seconds for other encodings.

\begin{figure}[ht]
    \centering
    \begin{subfigure}{0.45\columnwidth}
        \centering
        \includegraphics[width=\linewidth]{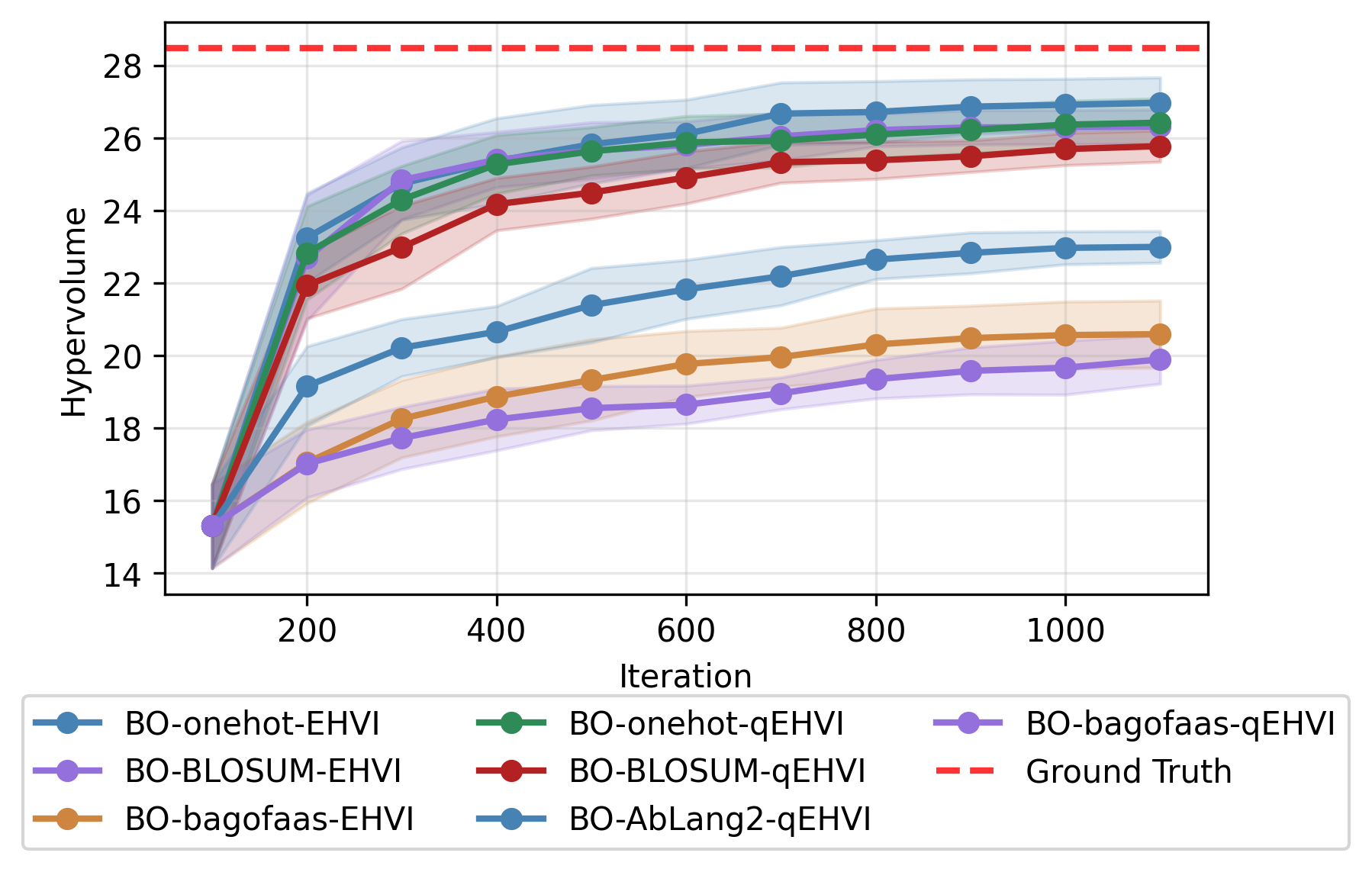}
        \caption{Comparisons of encodings for the EHVI and qEHVI (batch size 4) acquisition functions.}
        \label{fig:hypervolumecompare_encodings}
    \end{subfigure}
    \hspace{0.05\columnwidth}
    \begin{subfigure}{0.45\columnwidth}
        \centering
        \includegraphics[width=\linewidth]{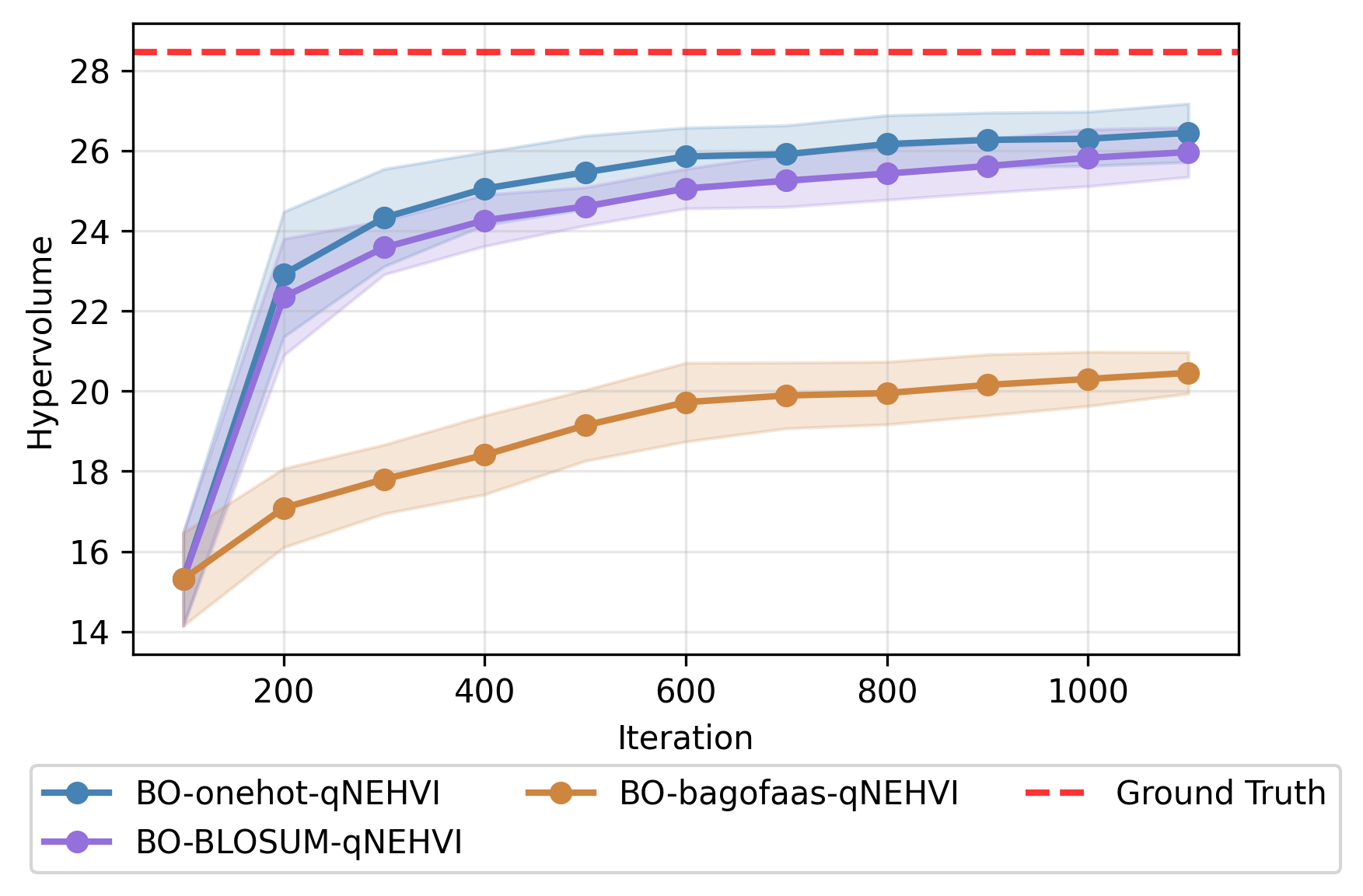}
        \caption{Comparisons of encodings for the qNEHVI acquisition function.}
        \label{fig:hypervolumecompare_encodingsnehvi}
    \end{subfigure}
    \caption{Plots comparing the hypervolume of the Pareto front by iteration of the BO algorithm for different encodings considered in Section~\ref{sec:vhh}.}
    \label{fig:hypcompare_enc}
\end{figure}

We saw in this example that the BLOSUM and one-hot encodings performed similarly. AbLang-2 actually performed worse than both BLOSUM and one-hot over 10 seeds, despite the encoding being more complex. Bag of amino acids was the worst performer of all the encodings. In further experiments, we used either BLOSUM or one-hot encoding, as both were relatively fast and performed equivalently well. 

\clearpage

\section{FULL COMPARISON FOR 5 MUTATIONS FOR ALL SEEDS}
\label{apx:allseeds}

\begin{figure}[h!]
    \centering
    \begin{subfigure}[b]{\columnwidth}
        \centering
        \includegraphics[width=0.85\columnwidth]{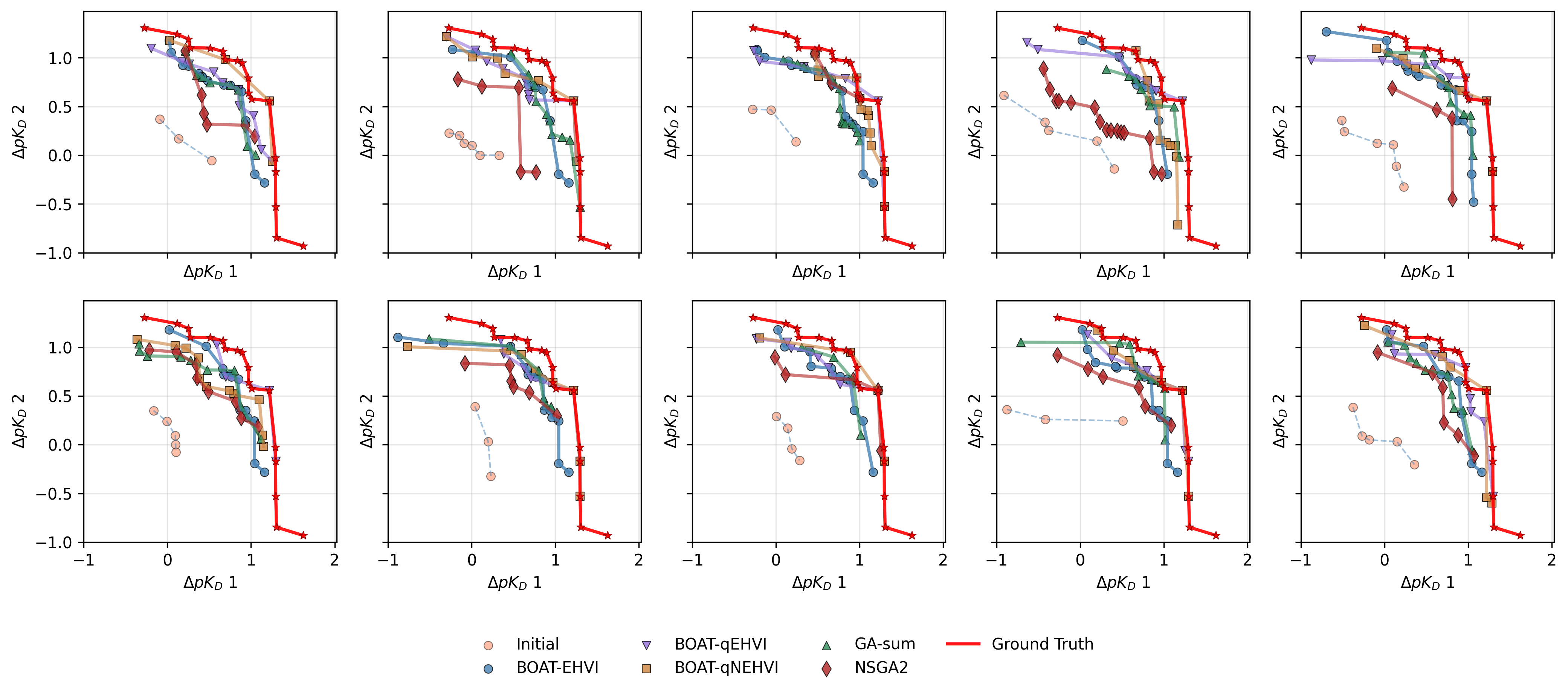}
        \caption{CDR1}
        \label{fig:fullparetocdr1}
    \end{subfigure}
    
    \begin{subfigure}[b]{\columnwidth}
        \centering
        \includegraphics[width=0.85\columnwidth]{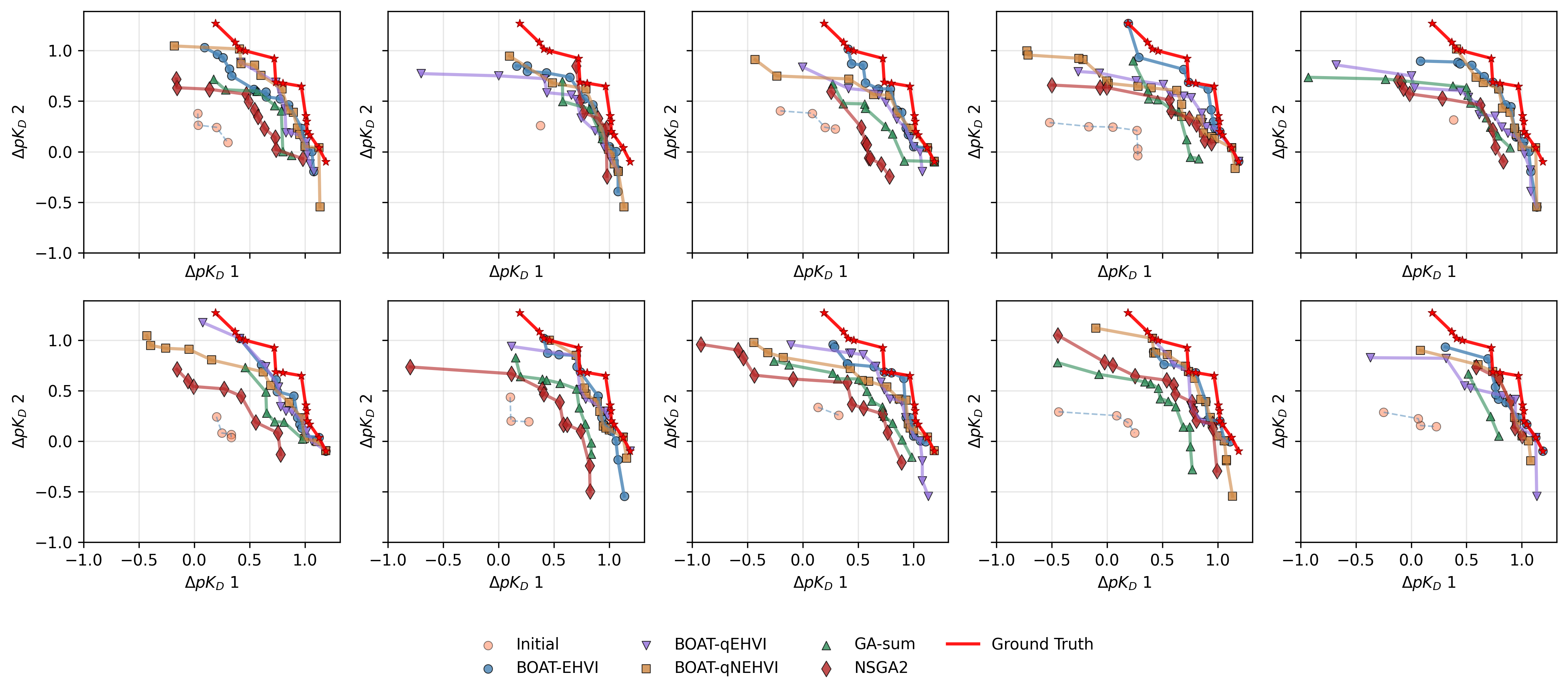}
        \caption{CDR2}
        \label{fig:fullparetocdr2}
    \end{subfigure}
    
    \begin{subfigure}[b]{\columnwidth}
        \centering
        \includegraphics[width=0.85\columnwidth]{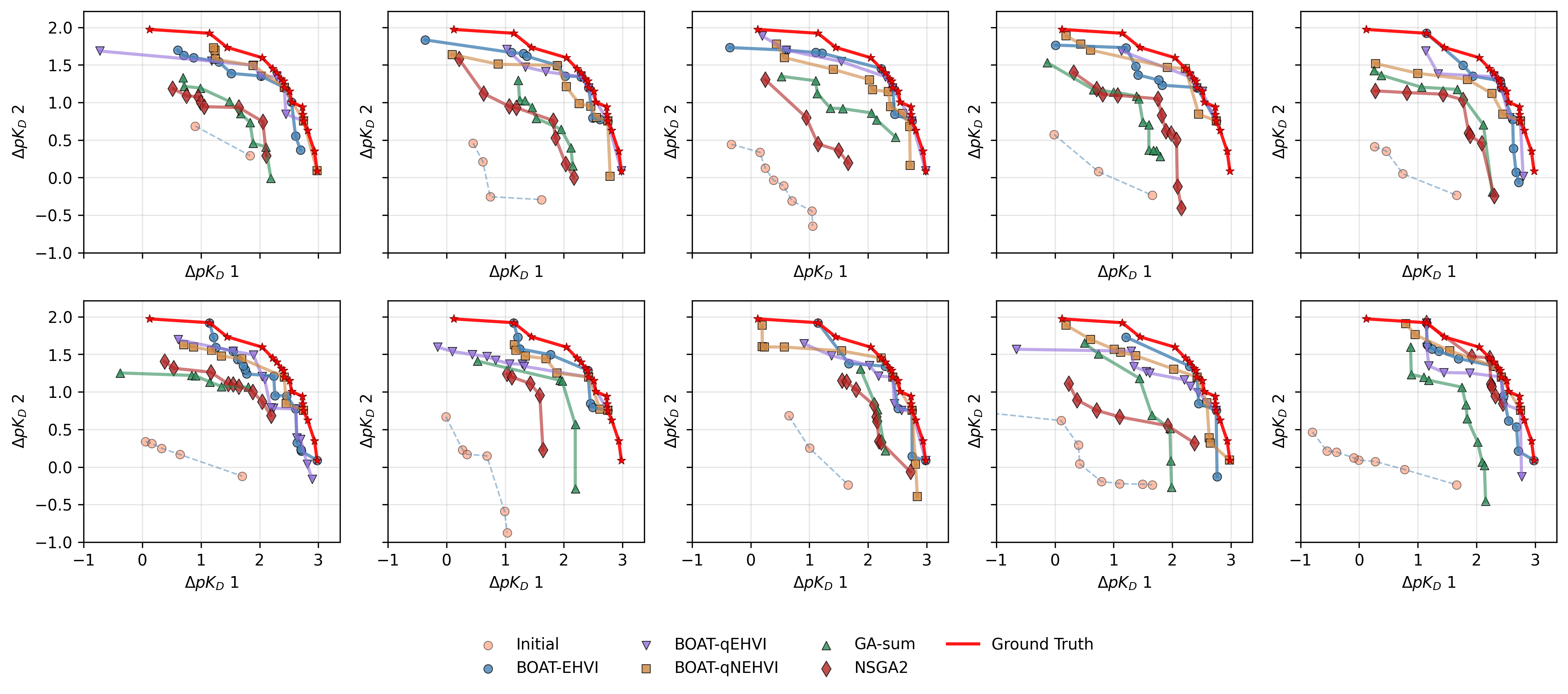}
        \caption{CDR3}
        \label{fig:fullparetocdr3}
    \end{subfigure}
  
    \caption{Plot comparing Pareto front by seed across different conditions for each CDR. All seeds are shown. It is clear especially in CDR3 runs that GAs tend to explore smaller Pareto fronts.}
    \label{fig:full5mutseeds}
\end{figure}

\begin{figure}[h!]
    \centering
    \begin{subfigure}[b]{\columnwidth}
        \centering
        \includegraphics[width=0.85\columnwidth]{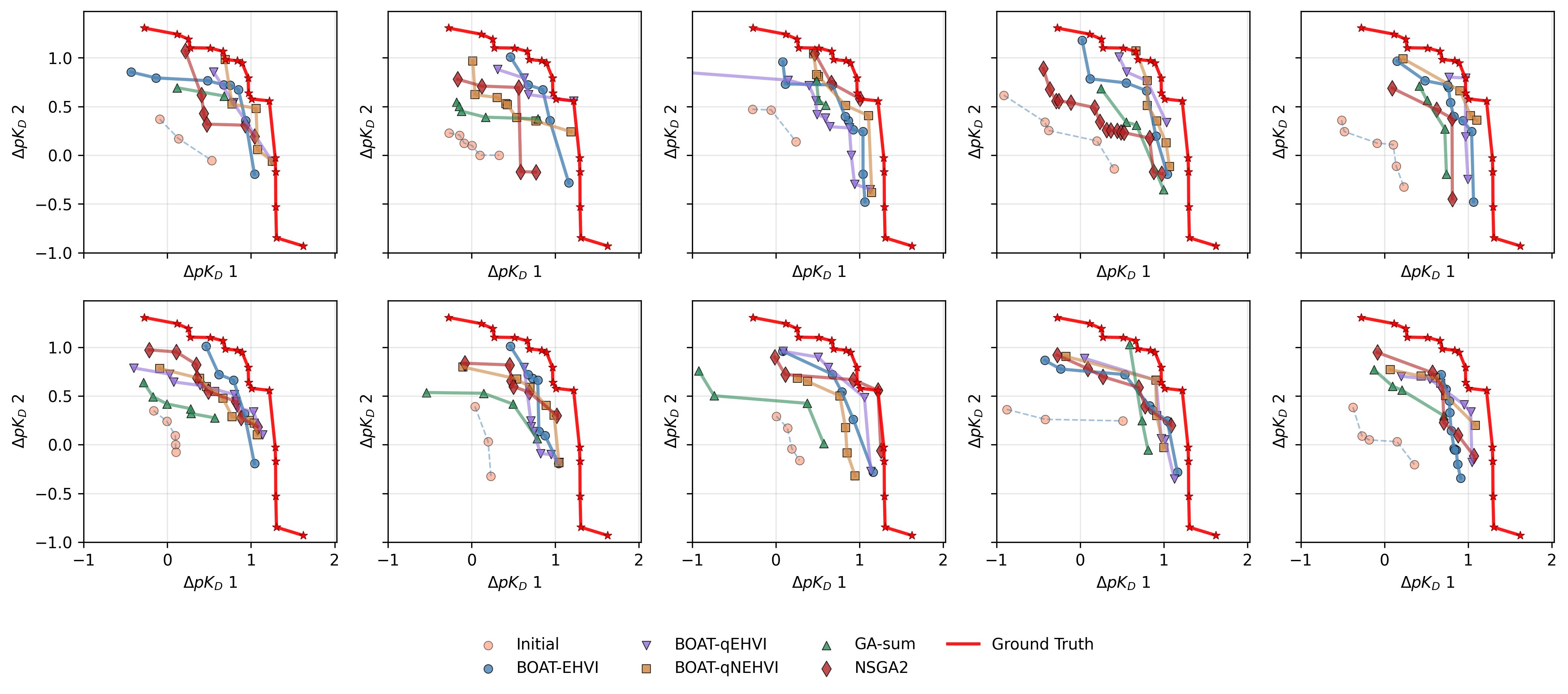}
        \caption{CDR1 (200 oracle calls)}
        \label{fig:fullparetocdr1300}
    \end{subfigure}
    
    \begin{subfigure}[b]{\columnwidth}
        \centering
        \includegraphics[width=0.85\columnwidth]{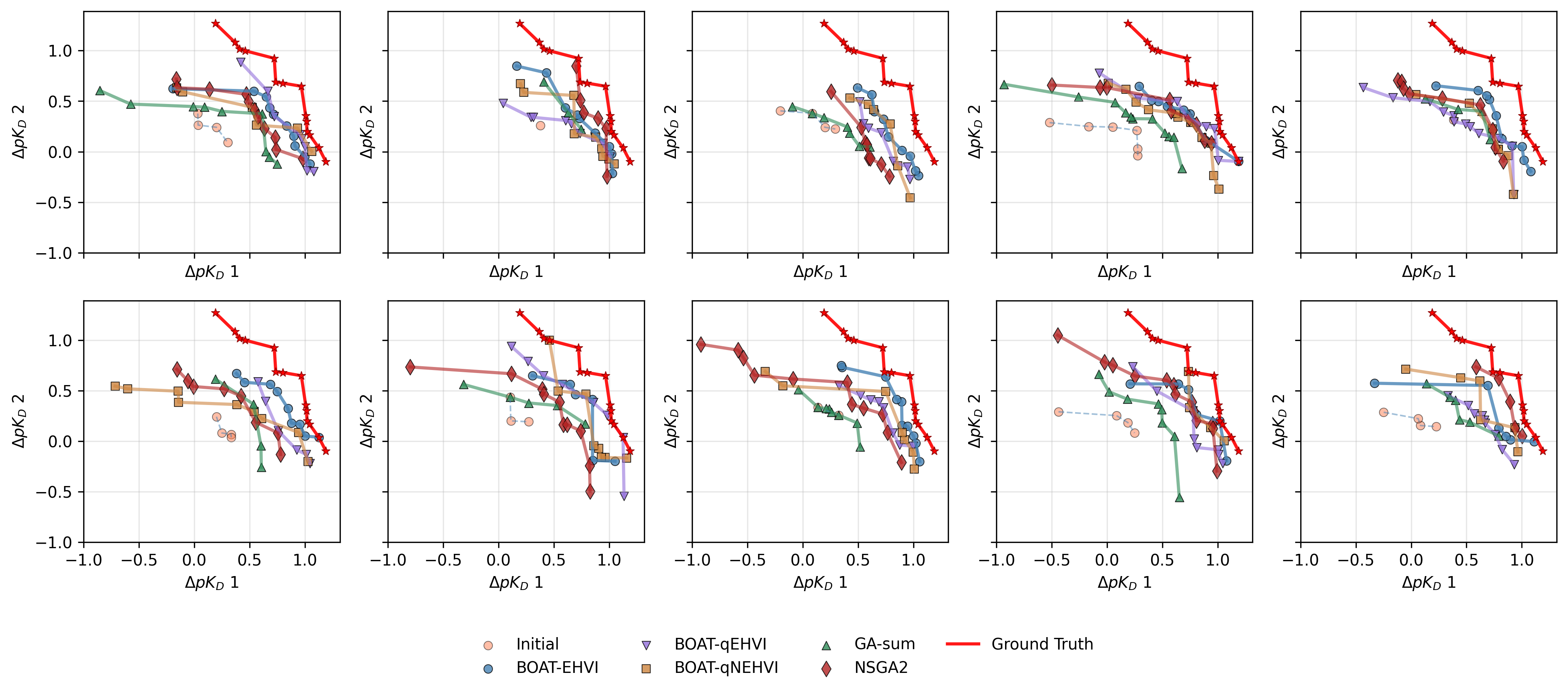}
        \caption{CDR2 (200 oracle calls)}
        \label{fig:fullparetocdr2300}
    \end{subfigure}
    
    \begin{subfigure}[b]{\columnwidth}
        \centering
        \includegraphics[width=0.85\columnwidth]{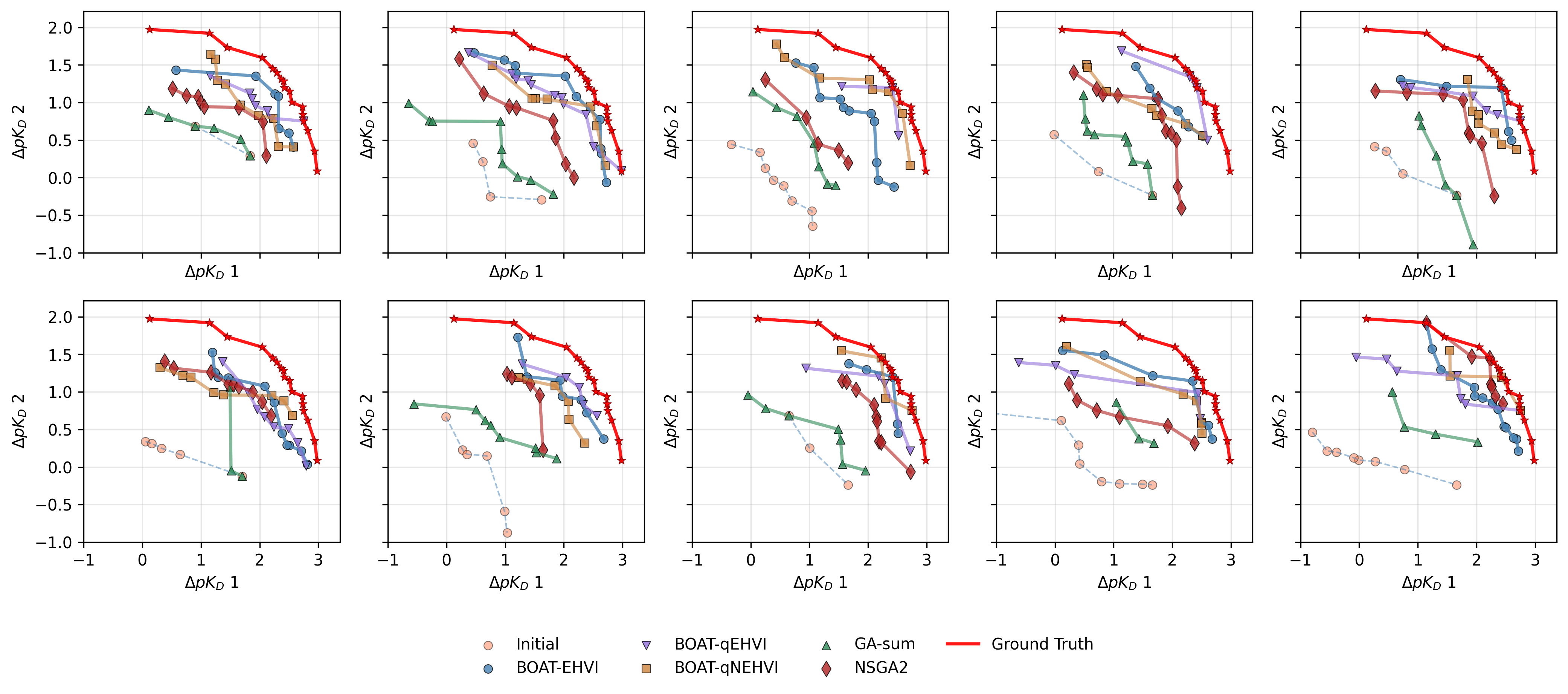}
        \caption{CDR3 (200 oracle calls)}
        \label{fig:fullparetocdr3300}
    \end{subfigure}
  
    \caption{Plot comparing Pareto front after 300 generated sequences (200 oracle calls after 100 initial sequences) by seed across different conditions for each CDR. All seeds are shown. Early in the algorithm (especially for CDR3), BOAT is able to explore much closer to the ground truth Pareto front compared to GAs.}
    \label{fig:full5mutseeds300}
\end{figure}

\clearpage

\section{COMPARING BATCH SIZE}
\label{apx:batch_size}

\begin{figure}[ht]
    \centering
    \begin{subfigure}{0.45\columnwidth}
        \centering
        \includegraphics[width=\linewidth]{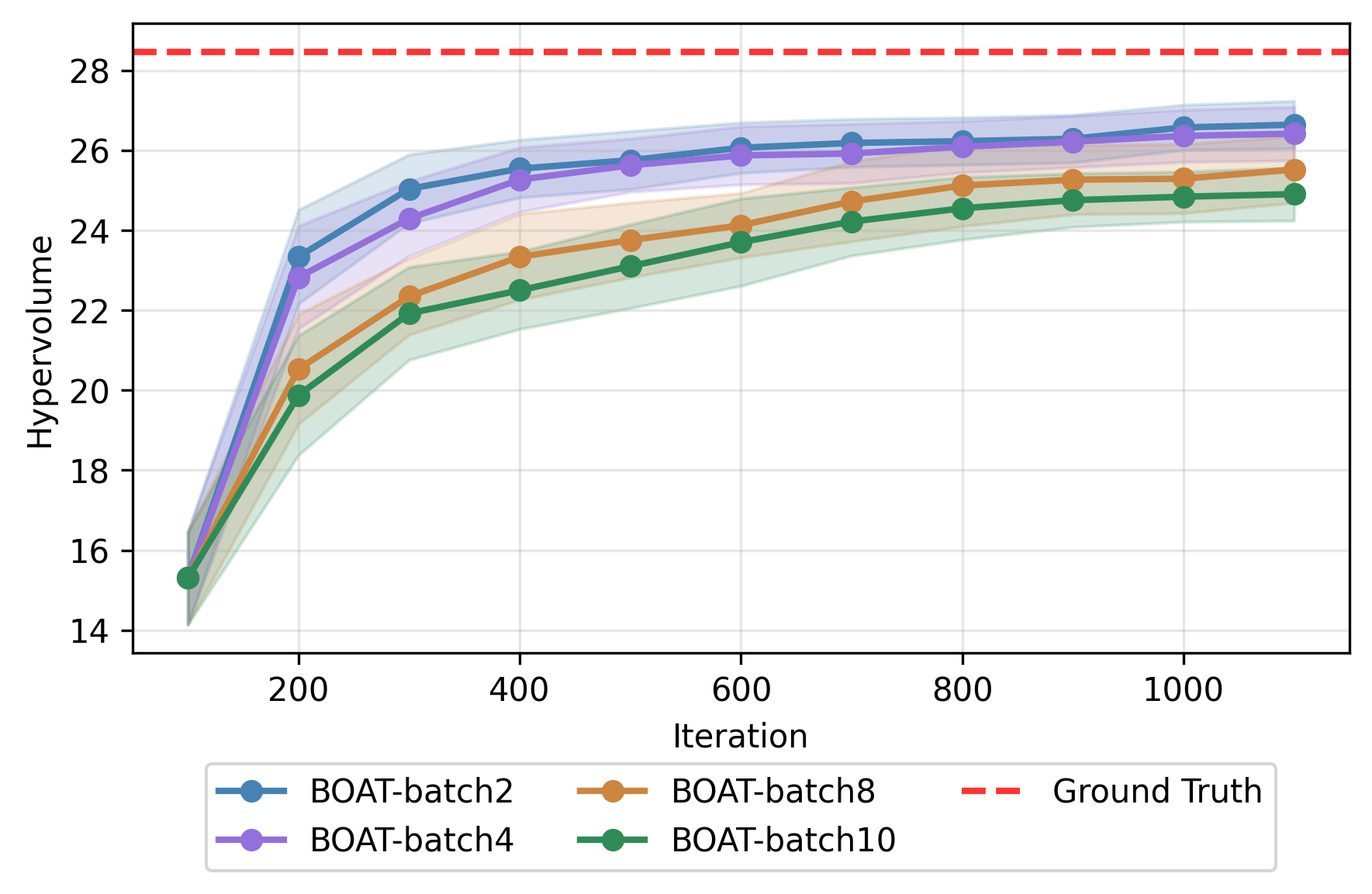}
        \caption{Hypervolume evolution comparison of batch sizes for CDR3, using qEHVI acquisition function.}
        \label{fig:hypervolumecompare_batchsize}
    \end{subfigure}
    \hspace{0.05\columnwidth}
    \begin{subfigure}{0.35\columnwidth}
        \centering
        \includegraphics[width=\linewidth]{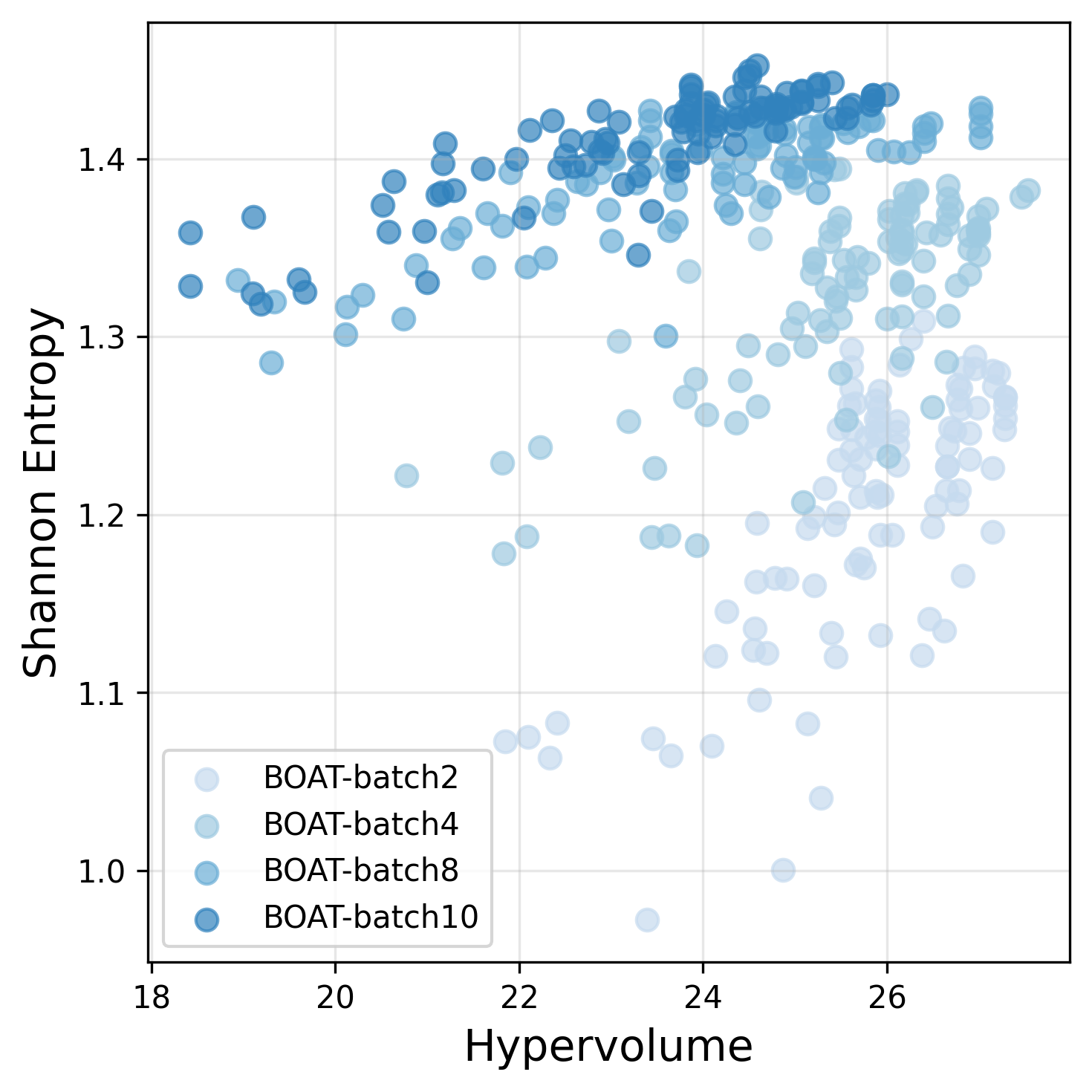}
        \caption{Scatter plots of hypervolume versus Shannon entropy for CDR3 optimization. Initial solutions are not shown.}
        \label{fig:shannon_batchsize}
    \end{subfigure}
    \caption{Plots for experiments when ablating batch size using the qEHVI acquisition function for CDR3 with 2 objectives and 5 maximum mutations, using batch sizes 2, 4, 8, 10. Other experimental details are kept the same as before. For (b), as before, each point represents the diversity and multi-objective performance of a population at a given optimization step, showing results for all seeds and batch sizes, and every 100 iterations.}
    \label{fig:compare_batchsize}
\end{figure}

We see in Figure~\ref{fig:hypervolumecompare_batchsize} that although we see decreased performance in terms of hypervolume improvement as we increase the batch size for the qEHVI acquisition function - especially towards the beginning of the algorithm - as we perform more iterations, results become more similar, with batch sizes 2 and 4 almost indistinguishable. As expected, we see more diversity as measured by Shannon entropy when using larger batch sizes, as seen in Figure~\ref{fig:shannon_batchsize}.

\section{4 MAXIMUM MUTATIONS}
\label{apx:4muts}

\begin{figure}[h!]
    \centering
    \begin{subfigure}[b]{\columnwidth}
        \centering
        \includegraphics[width=0.85\columnwidth]{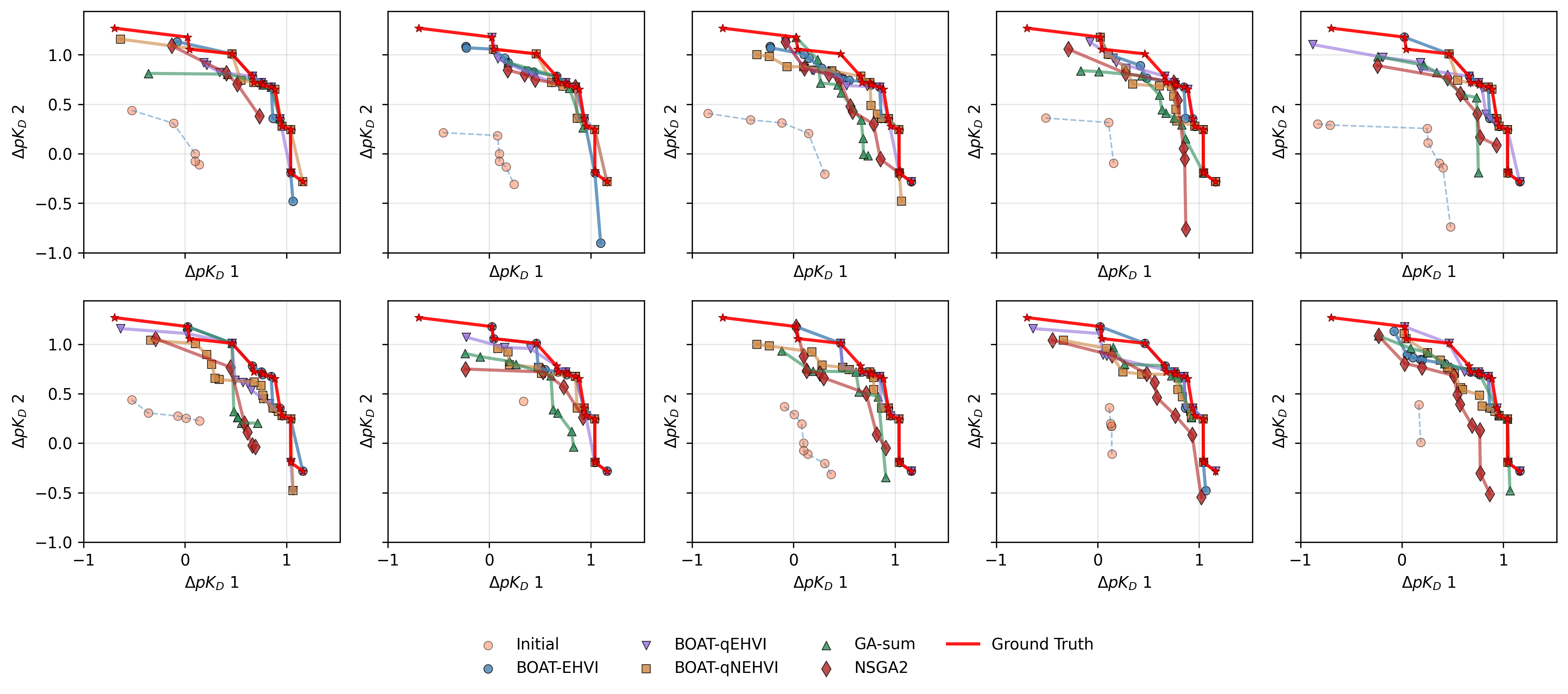}
        \caption{CDR1}
        \label{fig:fullparetocdr1_4mut}
    \end{subfigure}
    
    \begin{subfigure}[b]{\columnwidth}
        \centering
        \includegraphics[width=0.85\columnwidth]{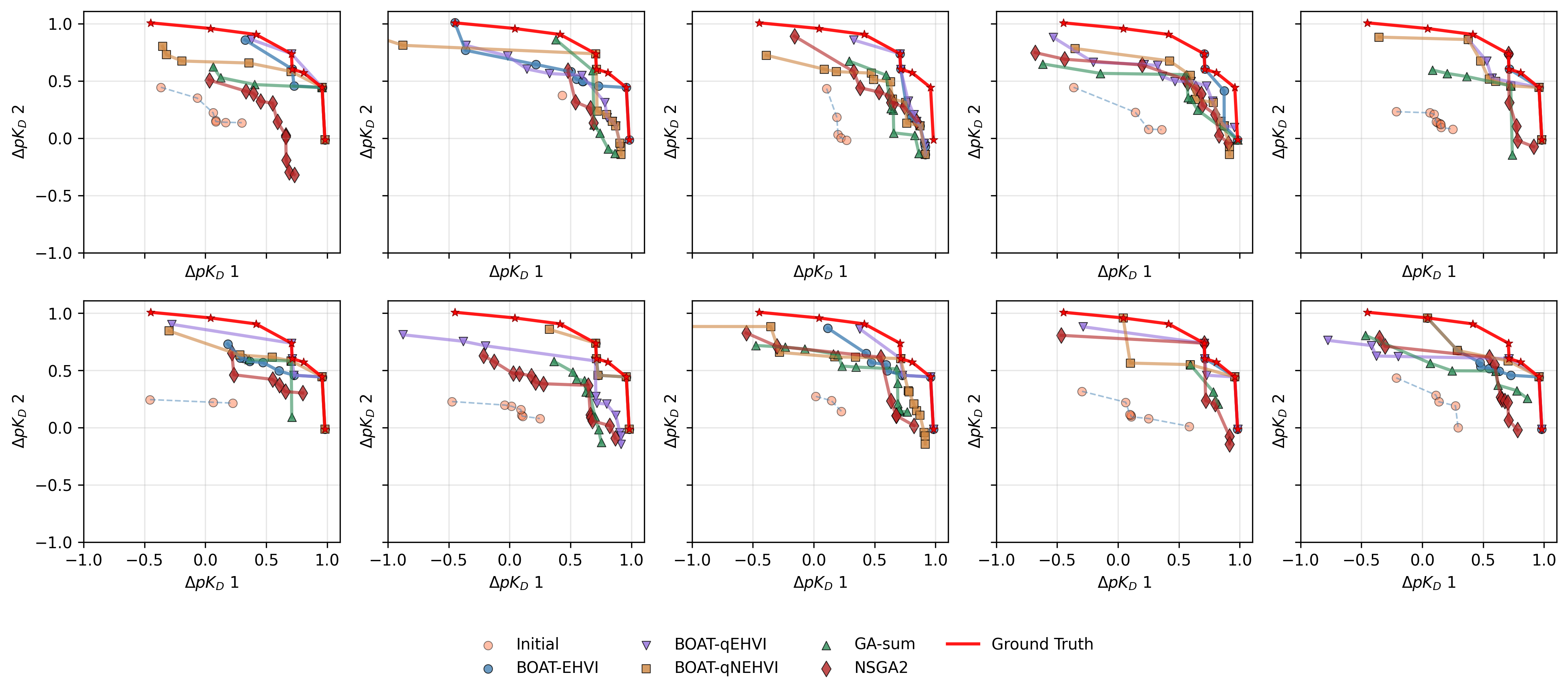}
        \caption{CDR2}
        \label{fig:fullparetocdr2_4mut}
    \end{subfigure}
    
    \begin{subfigure}[b]{\columnwidth}
        \centering
        \includegraphics[width=0.85\columnwidth]{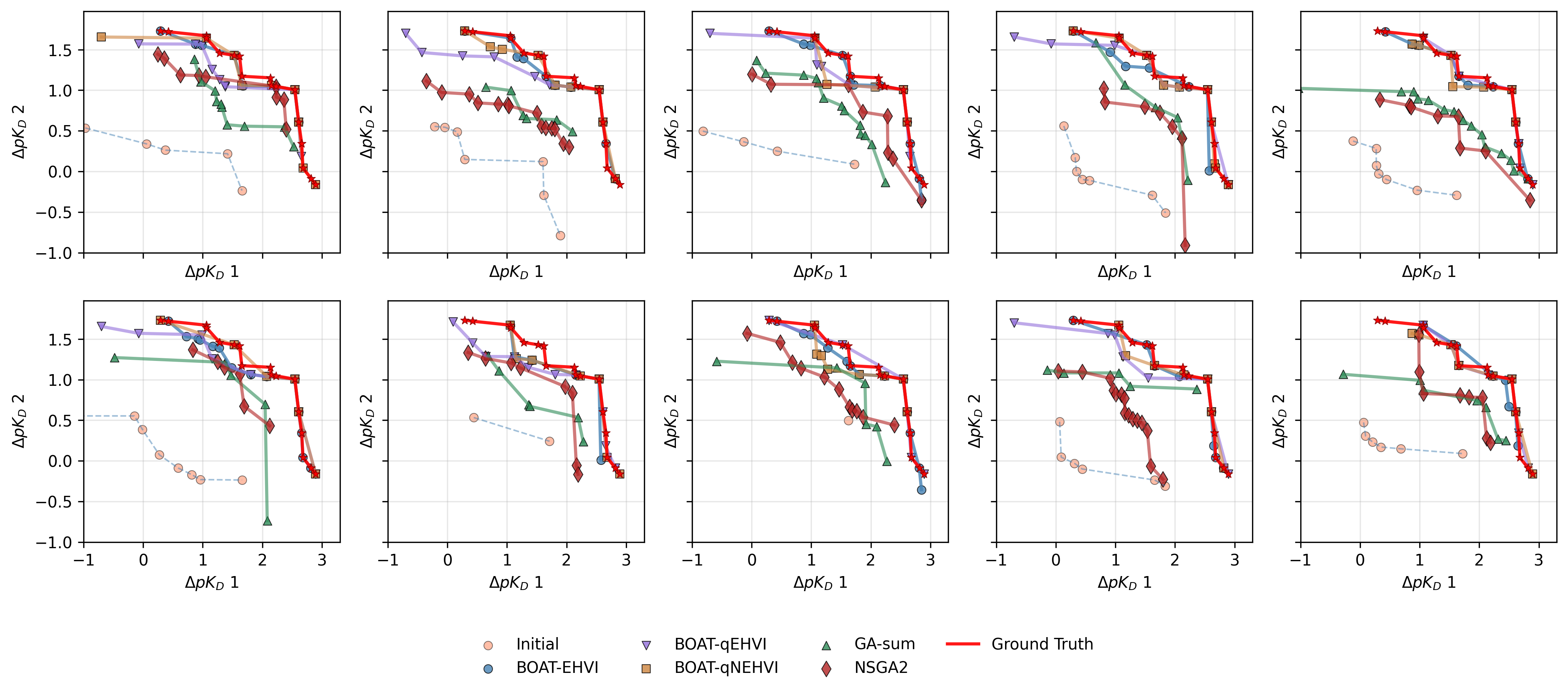}
        \caption{CDR3}
        \label{fig:fullparetocdr3_4mut}
    \end{subfigure}
  
    \caption{Plot comparing Pareto front by seed across different conditions for each CDR. All seeds are shown. Similarly to the 5 mutations example, GAs visually do not explore as much of the ground truth Pareto front compared to \method.}
    \label{fig:full4mutseeds}
\end{figure}

We ran a similar experiment to that in the main paper for 5 maximum mutations, but instead with 4 maximum mutations, which had a much smaller search space - CDR3 had 3975741 ground truth sequences, for example. We visualize the Pareto fronts found versus the ground truth in Figure~\ref{fig:full4mutseeds}.

\clearpage

\clearpage

\section{COMPARISON WITH LaMBO-2}
\label{apx:lambo}

\subsection{Details on the experimental setup}
\label{apx:lambo_exp_details}
Training of the discriminative head for LaMBO-2 was run for 100 epochs. For longer training, we observed the validation error increase. 561 sequences were held out for testing. We optimized 16 seed antibodies, which were all set to be the wild-type sequence, and ran LaMBO for 8 steps (with 4 guidance updates per step), leading to 256 total sequences being generated and evaluated. All other parameters followed default settings from the $\textit{cortex}$ GitHub implementation of LaMBO-2. For numerical stability and to phrase the optimization problem as an unconstrained maximization problem, we take the negative logarithm of the measured $k_D$ values and log-transform the expression data.

To provide \method~with a fair comparison, we used LaMBO-2's trained discriminative head $f^\ast$ as the black-box objectives for \method~(predicting affinity and expression). We initialized with 16 sequences containing up to 2 mutations from the wild-type, scored these with $f^\ast$, then ran \method~for 256 iterations in the sequential setting using EHVI, and for 64 iterations with a batch size of 4, i.e., also 256 oracle calls, for qEHVI. Given the modest training dataset size and corresponding uncertainty in the discriminative head's predictions, we also ran qNEHVI to explicitly model the noise in objective function evaluations with the same batch size as qEHVI. We ran both models with 5 different seeds. In order to maintain comparability across seeds, we trained the discriminative head once and then varied seeds across LaMBO and \method~optimization runs.
To address the naturalness of the sequences which is in-built for LaMBO-2, we explore two settings for \method, 1) a 3-objective setting in which we include ESM-2 as a third objective, 2) not including any naturalness constraint, i.e., only considering the given affinity and expression oracles.

\subsection{Additional results}
\label{apx:lambo_results}

\subsubsection{With naturalness constraint on \method}
\label{sec:lambo_w_plm}
\begin{figure}[ht]
    \centerline{\includegraphics[width=\textwidth]{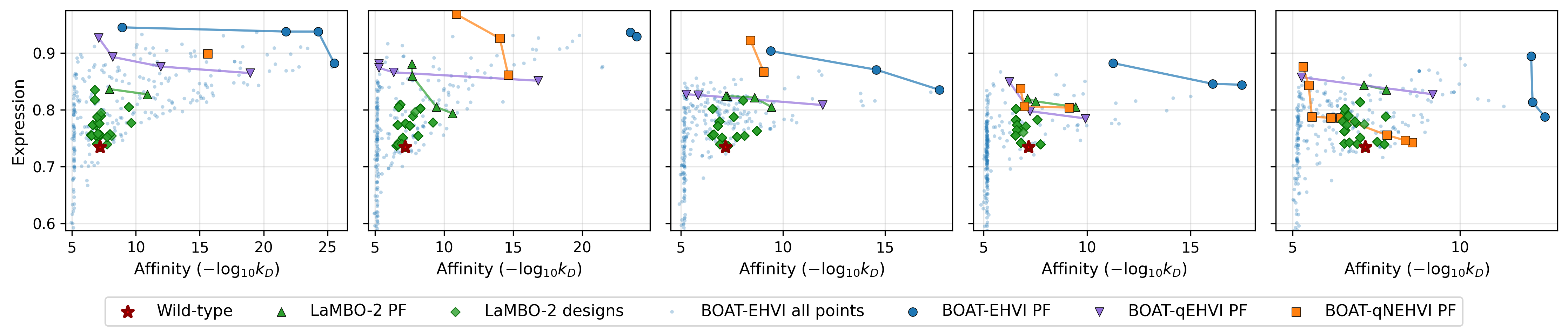}}
    \caption{Multi-seed comparison of \method~and LaMBO showing final Pareto fronts for \method-EHVI, \method-qEHVI, \method-qNEHVI, and all LaMBO designs across 5 optimization runs. While LaMBO got to optimize the 2 objectives for affinity and expression, the \method~variants additionally optimized for naturalness (ESM-2 likelihood). As the hypervolumes are fully comparable across seeds, we sorted them by descending terminal hypervolume for each method.} \label{fig:lamboseeds}
\end{figure}

The inclusion of a PLM likelihood can be seen as a "regularizer" on sequences, as sequences that are more likely to be found in nature will receive higher scores.
We ran the optimization for \method-EHVI, \method-qEHVI, and \method-qNEHVI. However, in the 3-objective setting, the qNEHVI is too slow to be recommended for practical applications.
For the 5 seeds tested, we plot the final Pareto front for the \method~variants, all LaMBO designs, the original wild-type point (with its predicted affinity and expression), as well as the designs found by \method-EHVI in Figure~\ref{fig:lamboseeds}.
These are the designs corresponding to the final state of the hypervolume progression plotted in Figure~\ref{fig:lambo_hypervolume}.

Figure~\ref{fig:lamboseeds} demonstrates the capability of all \method~methods to push the Pareto front, while LaMBO-2 designs exhibit a less explorative behaviour and produce sequences that are more similar to the parent. \method~explores a lot more aggressively than LaMBO-2 by mutating further away from the parental at a faster rate.
The plots also illustrate issues of the discriminative head, which predicts unrealistically large values for affinity for some sequences. It can be further seen that many of the \method-generated designs seem to be close to an affinity cut-off at 5. This is due to the dataset containing many sequences with $-\log_{10} k_D = 5$, which is likely the value assigned to all sequences without measurable binding. Remarkably, \method-EHVI finds the sequences with unrealistically low predicted $k_D$ values across seeds, while the batch versions do not. We postulate that the combinatorially larger amount of candidate batches in the batch versions makes it less likely to select one a batch containing one of these extreme but rare sequences than in the sequential case.

\subsubsection{Without naturalness constraint on \method}

\begin{wrapfigure}{r}{0.3\textwidth}
    \vspace{-8pt}  
    \centerline{\includegraphics[width=0.3\textwidth]{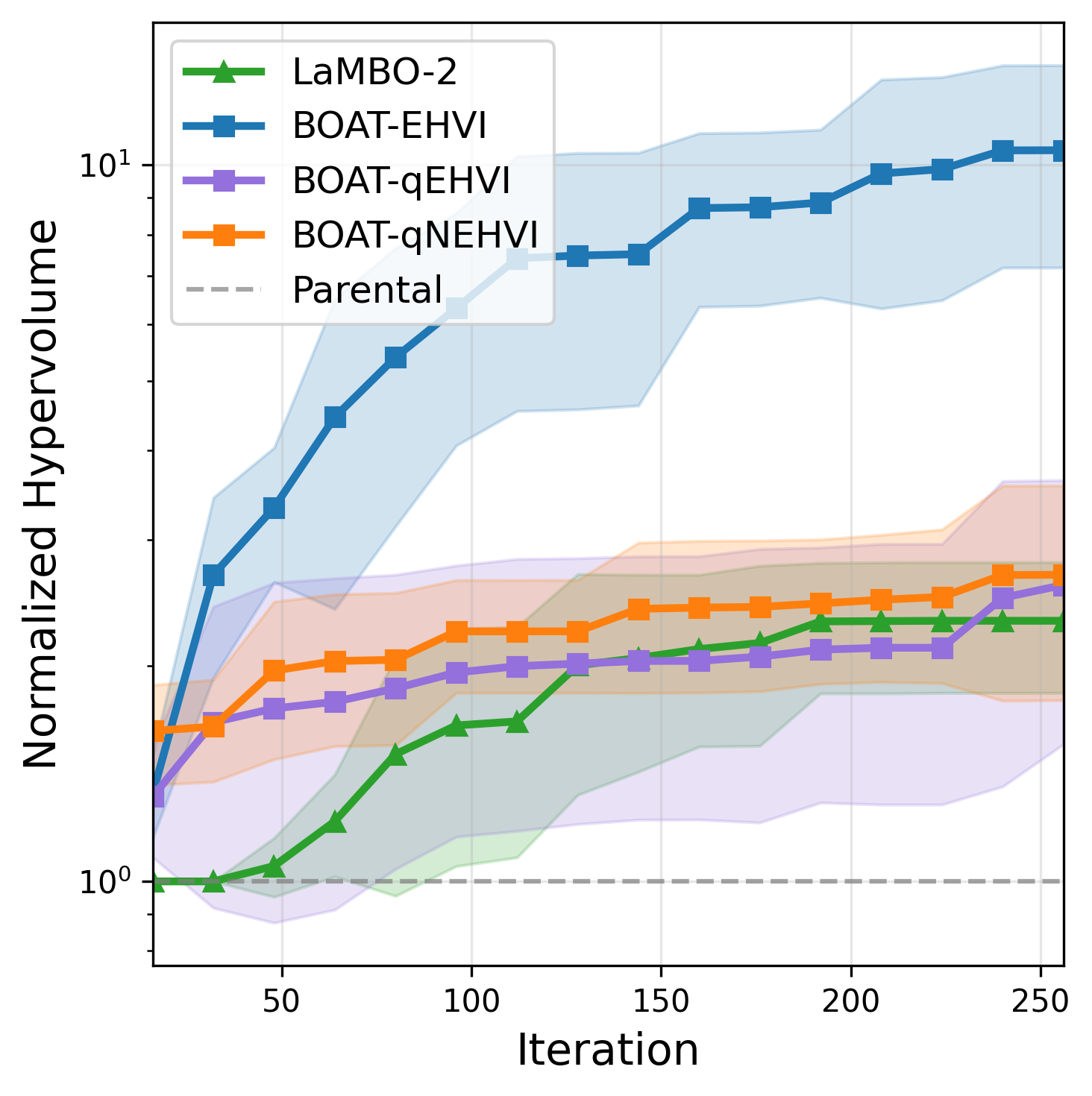}}
    \vspace{-5pt}   
    \caption{Hypervolume progression for LaMBO-2 and \method~variants when omitting naturalness constraints on \method.} \label{fig:lambo_hv_wo_plm}
    \vspace{-8pt}  
\end{wrapfigure}

When omitting naturalness constraints on \method, it becomes obvious in Figure~\ref{fig:lambo_hv_wo_plm} that \method~variants aggressively push the hypervolume by evaluating sequences with unrealistic affinity values caused by overfitting in the discriminative head.
Yet, this highlights that \method~in principle has the capability of finding such interesting sequences in more trustworthy oracles. LaMBO-2 is instead more conservative in optimizing leads, which prevents it from falling into the pitfall of proposing out-of-distribution sequences, but also keeps it from exploring more diverse sequences.

\begin{figure}[h]
    \centerline{\includegraphics[width=\textwidth]{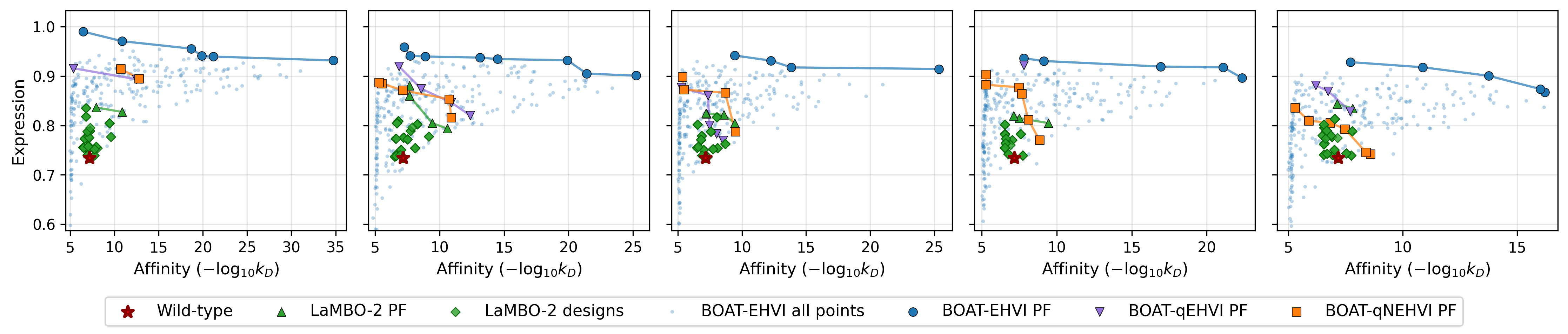}}
    \caption{Multi-seed comparison of \method~and LaMBO showing final Pareto fronts for \method-EHVI, \method-qEHVI, \method-qNEHVI, and all LaMBO designs across 5 optimization runs, where both LaMBO and \method~got to optimize the two objectives of affinity and expression predicted by LaMBO's discriminative head. As the hypervolumes are fully comparable across seeds, we sorted them by descending terminal hypervolume for each method.} 
    \label{fig:lamboseeds_wo_plm}
\end{figure}

Figure~\ref{fig:lamboseeds_wo_plm} shows the Pareto fronts corresponding to the final hypervolumes in Figure~\ref{fig:lambo_hv_wo_plm}. Even more than in the naturalness-constrained setting discussed in Section~\ref{sec:lambo_w_plm}, we observe the sequential version of \method~to discover sequences where the discriminative head fails.

The obvious quality issues of LaMBO-2's discriminative head underlines the benefit of the modular design of \method, which permits including tailored oracles for particular properties. While LaMBO-2 can be seen as an elegant lab-in-the-loop approach which requires little human intervention, the limitation to a fixed architecture for the discriminative head can also be seen as a defect that impedes leveraging auxiliary information that is not present in the data used for training. Not only does LaMBO-2 not overcome the need for human oversight and we observed, it requires profound technical understanding to perform low-level adaptations on the model to given task. With the possibility to interface externally built and validated oracles, we claim that \method~does not require the same depth of technical understanding to run.

